\documentclass[final]{siamonline190516}

\usepackage{enumitem}
\usepackage{makecell}
\usepackage{booktabs}
\usepackage{appendix}

\usepackage[latin1]{inputenc} % Definition des Eingabeencodings, also wie Umlaute umgesetzt werden
\usepackage[T1]{fontenc}      % Zeichensatzkodierung T1
\usepackage{ae}               % Verwendung von AE-Fonts
\usepackage{amsmath,amstext,amsfonts,amssymb,latexsym}
\usepackage{xfrac}

\usepackage{todonotes}

\usepackage{graphicx}         % for using includegraphics
\usepackage{epsfig}           % for using epsfig <-- uses includegraphics
\usepackage{caption,subcaption} % http://en.wikibooks.org/wiki/LaTeX/Floats,_Figures_and_Captions

\usepackage{enumitem}

\usepackage{pgfplots}
\usepackage{tikz}
\usetikzlibrary{calc,trees,positioning,arrows,chains,shapes.geometric, arrows.meta, decorations.pathreplacing,decorations.markings,decorations.pathmorphing,shapes,matrix,shapes.symbols,positioning,scopes,backgrounds}

\tikzset{join/.code=\tikzset{after node path={%
\ifx\tikzchainprevious\pgfutil@empty\else(\tikzchainprevious)%
edge[every join]#1(\tikzchaincurrent)\fi}}}

\tikzset{>=stealth',every on chain/.append style={join},
         every join/.style={->}}
\tikzstyle{labeled}=[execute at begin node=$\scriptstyle,
   execute at end node=$]

\renewcommand{\phi}{\varphi}

\newcommand{\op}{\text{op}~}
\newcommand{\enc}{\text{enc}}
\newcommand{\dec}{\text{dec}}

\newcommand{\encx}{_\enc^x}
\newcommand{\decx}{_\dec^x}
\newcommand{\ency}{_\enc^y}
\newcommand{\decy}{_\dec^y}
\newcommand{\aee}{^\text{\tiny{AE}}}
\newcommand{\sig}{^\text{\scalebox{0.8}{$\Sigma$}}}

\newcommand{\R}{\mathbb{R}}

\newcommand{\N}{N}
\newcommand{\NN}{\mathcal{N}}

\newcommand{\A}{\mathcal{A}}

\renewcommand{\L}[1]{\text{L}^{#1}(\Omega)}

\newcommand{\X}{\mathcal{X}}%
\newcommand{\Y}{\mathcal{Y}}%
\newcommand{\XX}{X}%
\newcommand{\YY}{Y}%
\newcommand{\ZZ}{Z}%
\renewcommand{\L}{L}%
\newcommand{\Id}{\textnormal{Id}}

\DeclareMathOperator*{\argmax}{\textnormal{argmax}}%
\DeclareMathOperator*{\argmin}{\textnormal{argmin}}

\renewcommand{\tilde}{\widetilde}

\newcommand{\norm}[1]{\| #1 \|}

\newcommand{\dd}{\partial}%
\newcommand{\dId}{\delta\hspace{0.3mm}\Id}
\newcommand{\Vn}{\scriptscriptstyle V_n}

\graphicspath{{images}}
\usepackage[export]{adjustbox}

\usepackage[a4paper, left=2.7cm, right=2.7cm, top=2.9cm, bottom=3.3cm]{geometry} %use in case of ArXiV version (smaller margins, centred, figure on first page)

\usepackage{changepage}

\definecolor{nice_green}{RGB}{28, 130, 35}
\definecolor{nice_purple}{RGB}{170, 35, 140}
\definecolor{middlebluee}{RGB}{0,95,155}
\definecolor{redd}{RGB}{195,60,60}
\definecolor{darkgreenn}{RGB}{80,144,80}

\title{Learned SVD: solving inverse problems via hybrid autoencoding}

\author{Yoeri E. Boink\footnotemark[2] \thanks{Multi-Modality Medical Imaging and Biomedical Photonic Imaging, Technical Medical Centre, University of Twente, NL (\email{y.e.boink@utwente.nl})}
\and Christoph Brune\thanks{Department of Applied Mathematics, University of Twente, NL (\email{y.e.boink@utwente.nl, c.brune@utwente.nl})}}

\begin{document}

\maketitle

\begin{abstract}
Our world is full of physics-driven data where effective mappings between data manifolds are desired. There is an increasing demand for understanding combined model-based and data-driven methods. We propose a nonlinear, learned singular value decomposition (L-SVD), which combines autoencoders that simultaneously learn and connect latent codes for desired signals and given measurements. We provide a convergence analysis for a specifically structured L-SVD that acts as a regularisation method. In a more general setting, we investigate the topic of model reduction via data dimensionality reduction to obtain a regularised inversion. We present a promising direction for solving inverse problems in cases where the underlying physics are not fully understood or have very complex behaviour. We show that the building blocks of learned inversion maps can be obtained automatically, with improved performance upon classical methods and better interpretability than black-box methods.
\end{abstract}


\section{Introduction}\label{sec:intro}%
% Learned SVD: Solving inverse problems by deep learning autoencoders
We are living in a world full of physics-driven data with an increasing demand for combining \textit{model-based} and \textit{data-driven} approaches in areas of science, industry and society. In many cases it is essential to reliably recover hidden multi-dimensional model parameters (signals) $x\in\XX$ from noisy indirect observations (measurements) $y^\delta\in\YY$, e.g. in imaging or sensing technology in medicine, engineering, astronomy or geophysics. These \textit{inverse problems}, $y^\delta = \A(x) + \eta^\delta$, are often ill-posed, suffering from non-uniqueness and instability in direct inversion. Classical model-based research on inverse problems has focused on variational regularisation methods to guarantee existence and stable approximation of solutions under uncertainty like noise $\eta^\delta$ in the measurements \cite{Engl1996, Benning2018}. For linear inverse problems the singular value decomposition (SVD) \cite{Golub1965} is a classical tool to directly construct a regularised inverse, e.g. in the sense of Tikhonov regularisation \cite{Engl1996}. Recent research in inverse problems has focused on combining deep learning with model-based approaches based on knowledge of the underlying physics \cite{Arridge2019}. Precise knowledge is often not available; for now we rely mainly on empirical evidence that such approaches can still be applied when one makes use of inexact operators that approximate the exact physical process \cite{Hauptmann2018b}. The main limitation of such methods are that they require an iterative application of expensive, possibly nonlinear mappings. Moreover, they are hard to interpret due to the lack of connection between data structure and structure of the mapping.

\vspace{-2mm}%
\begin{figure}[!ht]
    \centering
    \resizebox{0.81\linewidth}{!}{%
    \begin{tikzpicture}
    \input{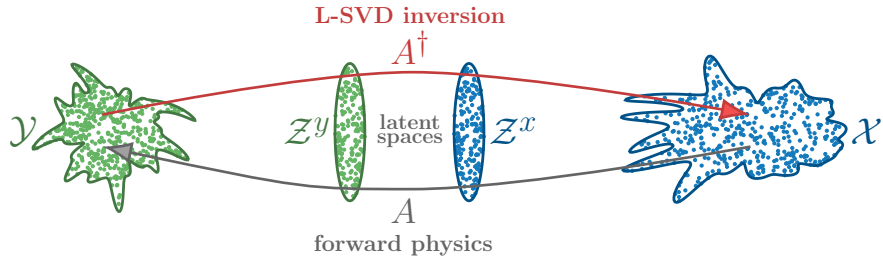}
    \end{tikzpicture}}
    \vspace*{-.5em}%
    \caption{L-SVD learns the inversion mapping via a hybrid nonlinear data manifold learning.}
    \label{fig:intro-tikz}%
\end{figure}
\vspace{-5mm}%
\newpage
We propose the `learned singular value decomposition' (L-SVD): a direct method that provides the inversion procedure with an explainable connection between measurements and signals. The method does not rely on an iterative application of expensive mappings; it does not need any information on the mapping at all. L-SVD makes use of two connected autoencoders: the first one encodes measurement $y^\delta$ to latent code $z_y$, while the second one encodes signal $x$ to latent code $z_x$ in a nonlinear way; both latent codes are connected with a linear `scaling' layer. The training of all parameters is done simultaneously, which enforces the latent codes to preserve as much information on the measurements and signals as possible, while making sure that the codes have very similar structure. After training, a reconstruction is obtained by consecutive application of encoding, scaling and decoding (see Figure \ref{fig:intro-tikz}).

In case a forward mapping is available, it can be used to extract an effective encoding and decoding, e.g. via the SVD. In such cases, the L-SVD can be shaped to a data-driven Tikhonov regularisation method, for which we prove convergence with respect to noise. In case a forward mapping is not available, L-SVD allows to learn the nonlinear inversion dynamics via two autoencoders and a linear scaling layer. This has the advantage that advances in interpretable autoencoding have a direct effect on the L-SVD method, making it easier to analyse than other fully learned inversion methods. Assuming that the autoencoders can be trained with high accuracy, finding the connection between both codes is a much lower dimensional problem than finding a nonlinear map between the measurements and signals directly. Moreover, in a semi-supervised scenario where not all training data is available in supervised pairs, the autoencoders still allow to learn from all samples, which is not possible in most supervised neural networks.

Current research in inverse problems is focused at developing theory for combining data-driven deep learning with model-based approaches \cite{Arridge2019}. Recently developed methods with theoretical guarantees such as convergence and stability include those in which a regularisation term is explicitly learned \cite{Lunz2018, Li2020} and those where an initial imperfect reconstruction is post-processed by a neural network \cite{Schwab2018a, Boink2020}. These methods are two-step procedures and require knowledge of the forward mapping, while L-SVD provides a one-step procedure without this knowledge. A series of works about the optimal regularised inverse matrix (ORIM) investigated the problem of finding a linear inverse matrix for a noisy inverse problem, both for the case where the linear forward matrix is known \cite{Chung2014, Chung2015} and for the case where it is not known \cite{Chung2014}, i.e. a fully learned scenario. Similar to the L-SVD, the idea of data-driven approximation of nonlinear mappings via model reduction was proposed in \cite{Bhattacharya2020}. In that paper, the interest is in approximating a forward mapping, while we are interested in solving an inverse problem. The idea of connecting two autoencoders was exploited in \cite{Zeng2017} and \cite{Gupta2017}, where the authors solved a superresolution and a deconvolution problem with a patch-based method. In our work, we consider a more general method that does not assume identical domains for measurement and signal. An extensive literature review of similar methods and the embedding of L-SVD in its research context is given in Section \ref{sec:literature}.

\subsection{Contribution}\label{sec:contrib}
This paper proposes the learned singular value decomposition (L-SVD), a general data-driven method that nonlinearly encodes (compresses) data in two vector spaces and connects them in an easy-to-understand way. The contributions of our method can be seen as the extension of existing methods in the following way:
\begin{enumerate}
%From solving inverse problem model-based to solving them data-driven:
\item \textbf{Data-driven solution of inverse problems:} L-SVD is a nonlinear generalisation of Tikhonov regularisation in Bayesian inverse problems \cite{Evans2002, Kaipio2005, Stuart2010} and piecewise linear estimates \cite{Yu2012}, which are linear data-driven variants of classical SVD approaches. We show that L-SVD can be shaped to a data-driven Tikhonov regularisation for which we provide a convergence analysis. In general, L-SVD requires no a-priori information on the forward mapping to achieve good reconstruction quality.
%From linear to nonlinear:
\item \textbf{Improved generalisation via nonlinear hybrid encoding:} Autoencoders show that nonlinear encoding provides better encodings than linear encoding \cite{Hinton2006}. L-SVD shows that this is also the case when encodings are used to solve inverse problems. An autoencoder can act as a regulariser when attached to a supervised neural network that is trained for a supervised task, e.g. classification \cite{Zhang2016, Le2018}. L-SVD makes use of two autoencoders for the task of solving an inverse problem, which gives improved generalisation performance. Moreover, it enables high-quality reconstruction in a semi-supervised setup.
\end{enumerate}
%\YB{@Christoph: I am not completely sure how we should write the `contribution' of this work to the field. Previously, we showed three ways in which it generalised %existing methods. Obviously, this was not really a contribution list. Right now I chose for a mix between the two: I mention how our method extends existing methods %and mention how they should be seen as contributions. I am not fully happy with this, but right now I don't know how to improve it. It would be great if you would %have a critical look at this section. If necessary, rewrite the whole section.\\
%However, I did refer to these two points later in the paper (in section 6 for instance). So if something is changed, that has to be changed too.}

\subsection{Overview of the paper}
Throughout the paper, we make use of the above two perspectives to show the advantage of using the L-SVD method for addressing the previously mentioned limitations. In Section \ref{sec:IP_SVD}, a brief overview of the classical SVD and inversion methods is given, which serves as motivation for the L-SVD method. Next, a precise definition of L-SVD is provided in Section \ref{sec:model}, accompanied with various architecture choices that exploit its potential. One of these choices results in a data-driven Tikhonov regularisation, while a second one results in a fully learned L-SVD method. In Section \ref{sec:analysis}, we analyse the L-SVD method by showing its connection with Bayesian inverse problems, by showing a convergence analysis of the data-driven Tikhonov regularisation and providing a stability and error estimate for the fully learned L-SVD method. After that, in Section \ref{sec:literature}, the connection of our work with several fields of research are discussed. In Section \ref{sec:implementation}, we explain simulation experiments that show the transition from non-learned to learned, linear to nonlinear, and single to hybrid encoding. Results are provided in Section \ref{sec:results}, where we visualise these transitions by looking at the latent space, the decoded representations of the latent space and the dependency of L-SVD on noise. The section is completed with a comparison between L-SVD and state-of-the art reconstruction methods applied on a biomedical CT data set. In Section \ref{sec:conclusion}, we conclude with some remarks and outlook for future work.

\section{Motivation: SVD and inversion methods}\label{sec:IP_SVD}
The motivation of our L-SVD method can be found in the application of classical SVD and its variants in inversion methods. In our work we consider the finite dimensional version of the equation introduced in Section \ref{sec:intro}. That is, we make use of the `first discretise, then optimise' approach. We define the inverse problem as
\begin{equation}\label{eq:IP}
y^\delta = \A(x) +\eta^\delta, 
\end{equation}
where we wish to reconstruct the signals $x\in\XX\subseteq \R^m$ from measurements $y^\delta\in\YY \subseteq \R^n$ corrupted by additive noise $\eta^\delta\sim\N(0,\delta\text{Id})$. Here $\XX$ and $\YY$ are Banach spaces. The mapping $\A:\XX\mapsto \YY$ is in general a nonlinear one. For this section however, we assume a linear operator that we call $A$. Any $A$ can be written in its singular value decomposition: $A = USV^*$, where $U\in\R^{n\times n}$ and $V\in\R^{m\times m}$ are unitary matrices and $S\in\R^{n\times m}$ is a diagonal matrix with nonnegative real numbers $s_i$ (singular values) on the diagonal. We now summarise well-known inversion methods that can be written as the application of an SVD \cite{Golub1965}.

\subsection{Maximum likelihood estimator (MLE)}\label{sec:MLE}
The MLE is defined via the maximisation over $x$ given measurements $y^\delta$ \cite{Stuart2010}. Its solution $x_\text{\tiny MLE}$ is obtained by applying the Moore-Penrose inverse $(A^*A)^{-1}A^*$ to the measurements $y^\delta$. 
\begin{align}\label{eq:MLE}
x_\text{\tiny MLE} 	:&= \argmax_x p(x|y^\delta) = \argmin_x\norm{Ax-y^\delta}_{\ell^2}^2\nonumber\\
					& = (A^*A)^{-1}A^*y^\delta = VS^{-1}U^*y^\delta.
\end{align}

\subsection{Tikhonov regularisation}\label{sec:Tikhonov}
Tikhonov regularisation is a method which puts a uniform variance prior on the desired solution $x$. It solves an $\alpha$-weighted minimisation problem that can be solved directly via its regularised Moore-Penrose inverse:
\begin{align}\label{eq:Tikhonov}
x_\alpha :&= \argmin_x\norm{Ax-y^\delta}^2_{\ell^2}+\alpha\norm{x}^2_{\ell^2},\nonumber\\
 &= (A^*A+\alpha\Id)^{-1}A^*y^\delta = V\underbrace{\big(S^2+\alpha\Id\big)^{-1}S}_{S^{-1}_\alpha}U^* y^\delta. 
\end{align}
The diagonal elements of $S^{-1}_\alpha$ are defined as $s_i/(s_i^2+\alpha)$, which means that for smaller scales $s_i$, the new inverse scale goes to zero as $\alpha$ gets larger. The optimal $\alpha$ depends on the type of noise; usually $\alpha$ increases with noise level $\delta$. 

For a specific model choice (see Section \ref{sec:variations}), L-SVD becomes a data-driven Tikhonov regularisation method, for which we provide a convergence analysis in Section \ref{sec:nonl_Tikh}.

\subsection{Truncated SVD}\label{sec:TSVD}
The best Frobenius approximation of $A$ with rank $r$ is given \cite{Eckart1936} by the truncated SVD (T-SVD):
\begin{align}\label{eq:thin_SVD}
A_r :&= U_rS_rV^*_r = \argmin_{\tilde{A}}\norm{A-\tilde{A}}_\text{Fro}~ \text{ s.t. rank}(\tilde{A})=r.
\end{align}
Here we made use of the `thin' representation, where $U_r\in\R^{n\times r}$ and $V_r\in\R^{m\times r}$ consist of the top $r$ rows of $U$ and $V$ respectively. $S_r\in\R^{r\times r}$ is a diagonal square matrix that consists of the largest $r$ singular values of $A$. With the thin representation, we lose one desirable property, namely that of unitary matrices: while $U_r^*U_r = \Id_r = V_r^*V_r$ still holds, generally $U_rU_r^* \neq \Id \neq V_rV_r^*$.

T-SVD can be applied in an inversion method instead of the standard SVD for noisy measurements $y^\delta$: when $s_i$ becomes small for $i$ large, noise is amplified by $1/s_i$ in \eqref{eq:MLE}. This problem is mitigated by solving $x_\text{trunc} := V_rS_r^{-1}U_r^*y^\delta$ instead. It has the additional benefit that the thin decomposition is smaller than the full SVD, requiring less memory and computation time.

\section{The learned singular value decomposition}\label{sec:model}
In this section, we provide the general L-SVD method for solving inverse problems. It aims to solve the inverse problem as defined in \eqref{eq:IP}, where the forward mapping $\A$ may be nonlinear. L-SVD can be seen as a nonlinear learned variant of the inversion methods in Section \ref{sec:IP_SVD}, where $U^*$ is replaced by a nonlinear encoder and $V$ by a nonlinear decoder.

\subsection{Model statement}\label{sec:model_statement}
\begin{figure}[!ht]
\centering
\resizebox{0.6\linewidth}{!}{%
\begin{tikzpicture}
\def\lay{2}
\def\x{0}
\def\xx{2.5}
\def\xxx{5}
\def\xxxx{7.5}
\def\y{0}
\def\yy{0.5}
\def\yyy{2.5}

\definecolor{middlebluee}{RGB}{0,95,155}
\definecolor{darkbluee}{RGB}{1,82,137}
\definecolor{greenn}{RGB}{100,180,100}
\definecolor{darkgreenn}{RGB}{80,144,80}
\definecolor{redd}{RGB}{195,60,60}
\definecolor{lightredd}{RGB}{245,80,80}
\definecolor{grayy}{RGB}{180,180,180}

\node (u1) at (\x,\y) {$\mathbf{\tilde{z}_x}$};
\node (u2) at (\xx,\y) {$\mathbf{x}$};
\node (u3) at (\xxx,\y) {$\mathbf{z_x^\text{\tiny \bf{AE}}}$};
\node (u4) at (\xxxx,\y) {$\mathbf{\hat{x}^\text{\tiny \bf{AE}}}$};
\node (uu3) at (\xxx,\yy) {$\mathbf{z_x^\text{\scalebox{0.8}{$\mathbf{\Sigma}$}}}$};
\node (uu4) at (\xxxx,\yy) {$\mathbf{\hat{x}^\text{\scalebox{0.8}{$\mathbf{\Sigma}$}}}$};
\node (f2) at (\xx,\yyy) {$\mathbf{y^\delta}$};
\node (f3) at (\xxx,\yyy) {$\mathbf{z_y}$};
\node (f4) at (\xxxx,\yyy) {$\mathbf{\hat{y}^\text{\tiny \bf{AE}}}$};

\draw[->,line width=1.5pt, draw=grayy] (u1) to node [below] {$\phi\decx(\cdot)$} (u2);
\draw[->,line width=1.5pt, draw=middlebluee] (u2) to node [below] {$\phi\encx(\cdot)$} (u3);
\draw[->,line width=1.5pt, draw=middlebluee] (u3) to node [below] {$\phi\decx(\cdot)$} (u4);
\draw[->,line width=1.5pt, draw=redd] (uu3) to node [above] {$\phi\decx(\cdot)$} (uu4);
\draw[->,line width=1.5pt, draw=greenn] (f2) to node [above] {$\phi\ency(\cdot)$} (f3);
\draw[dashed,line width=1.5pt, draw=redd] (f2) to node [above] {} (f3);
\draw[->,line width=1.5pt, draw=greenn] (f3) to node [above] {$\phi\decy(\cdot)$} (f4);
\draw[->,line width=1.5pt, draw=grayy] (u2) to node [left] {$A (\cdot)+\eta^\delta$} (f2);
\draw[->,line width=1.5pt, draw=redd] (f3) to node [right] {$\Sigma(\cdot)$} (uu3);
\end{tikzpicture}}
\caption{Schematic of the L-SVD method. {\color{darkgreenn} \textbf{Green}}: autoencoder for measurement $y^\delta$. {\color{middlebluee} \textbf{Blue}}: autoencoder for signal $x$. {\color{redd} \textbf{Red}}: reconstruction procedure. The standard network does not use the gray connections for training. Note that $\phi\decx$ is used multiple times, but has shared weights.}
\label{fig:L-SVD-tikz}
\vspace{-5mm}
\end{figure}
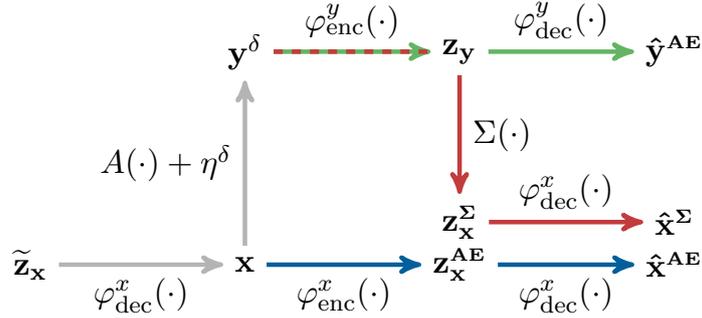

The L-SVD model (Figure \ref{fig:L-SVD-tikz}) is a trained neural network that consists of a measurement autoencoder (green), a signal autoencoder (blue) and a reconstruction component (red). Reconstruction $\hat{x}$ from measurement $y^\delta$ is obtained via the latent representations $z_x\in \ZZ^x\subseteq\R^k$ and $z_y\in \ZZ^y\subseteq\R^k$, which are part of the autoencoders. The latent space $\R^k$ is a low-dimensional space, i.e. $k\leq\min\{m,n\}$. A more formal definition is given as:
\vspace{4mm}
\begin{center}
\begin{adjustwidth}{0.04\textwidth}{0.04\textwidth} 
\begin{definition}\label{def:LSVD}
We define the nonlinear functions
$$\phi\ency:\YY\mapsto \ZZ^y,~~\phi\decy:\ZZ^y\mapsto \YY,~~\phi\encx:\XX\mapsto \ZZ^x,~~\phi\decx:\ZZ^x\mapsto \XX,~~\Sigma:\ZZ^y\mapsto \ZZ^x.$$
and the variables 
\begin{equation*}
\begin{split}
&z_y:=\phi\ency(y^\delta),\\ &\hat{y}\aee:=\phi\decy(z_y),
\end{split}
\quad\quad
\begin{split}
&z_x\aee:=\phi\encx(x),\\ &\hat{x}\aee:=\phi\decx(z_x\aee),
\end{split}
\quad\quad
\begin{split}
&z_x\sig:=\Sigma(z_y),\\ &\hat{x}\sig = \phi\decx(z_x\sig).
\end{split}
\end{equation*}
\end{definition}
\end{adjustwidth}
\vspace{4mm}
\end{center}

The L-SVD model is obtained by minimising a neural network loss function that consists of three parts:
\begin{equation}\label{eq:par_min}
\min_{\text{par}_\text{\tiny NN}}\left\{\sum_{i=1}^{\text{\#train}} \underbrace{D_1\big(\hat{x}_{(i)}\sig, x_{(i)}\big)}_{\text{reconstruction}} + \alpha_y \underbrace{D_2\big(\hat{y}_{(i)}\aee, y_{(i)}\big)}_{\text{autoencoder}} + \alpha_x \underbrace{D_3\big(\hat{x}_{(i)}\aee, x_{(i)}\big)}_{\text{autoencoder}}\right\},
\end{equation}
where we minimise over all trainable parameters par$_\text{\tiny NN}$ and over all samples $(i)$ in the training set. The distance functions $D_j(\cdot,\cdot)$ can be any metric; often used in neural networks are $\ell^2$, $\ell^1$ and $W^2$ (Wasserstein) metrics. The L-SVD model encodes measurements $y^\delta$ into a representation that contains sufficient information to approximately reconstruct the clean data $y$, while being able to map to an encoded representation that can approximately be decoded to the desired signal $x$. Since the output of the data autoencoder is the clean data $y$, instead of the corrupted measurements $y^\delta$, it can be seen as a denoising autoencoder (DAE) \cite{Vincent2010}. This means that noise will not necessarily be represented in the latent variable $z_y$. 

\subsubsection{Opportunities by various model choices}\label{sec:variations}
Below some specific model choices and variations on the standard model are discussed, which establish certain capabilities of the L-SVD model:
\begin{itemize}
\item \relax\emph{\textbf{Data-driven Tikhonov regularisation:}} With the SVD, a linear encoder and decoder can be derived from the operator $A$ (see Section \ref{sec:IP_SVD}). If this is combined with the nonlinear scaling function $\Sigma = \big(S^2+\alpha\NN (y^\delta)\big)^{-1}S,$ with $0<\NN(y^\delta)<\infty$ a nonlinear function, a data-driven Tikhonov functional is obtained. A convergence analysis is given in Section \ref{sec:nonl_Tikh}. 
\item \relax\emph{\textbf{Linearly connected nonlinear representations:}} Results from autoencoding \cite{Hinton2006} motivate the search for a nonlinear encoding and decoding such that a nonlinear representation of signals and measurements is obtained. The expressiveness of the nonlinear encoder and decoder allows us to restrict the scaling layer to be a square matrix $\Sigma\in\R^{k\times k}$, either full or diagonal. Generic stability and reconstruction estimates are given in Section \ref{sec:analysis_fully}. They depend on the reconstruction quality of the autoencoders, which can be freely chosen depending on the application at hand. This includes regularised autoencoders, such as sparse \cite{Ranzato2007} or contractive autoencoders \cite{Rifai2011} of a fully-connected or convolutional type.\newpage
\item \relax\emph{\textbf{Noise-aware $\Sigma$:}} The autoencoder on the measurement side can be chosen to be a regular autoencoder instead of a denoising autoencoder, with the effect that noise is represented in the encoded version $z_y$. This means that the latent dimension should be large enough, since unstructured noise can not be compressed. Moreover, this means that $\Sigma$ should be able to remove (part of) the noise, since the latent variable $z_x$ is noise-free.
\item \relax\emph{\textbf{Structured latent space:}} No specific structures of the latent spaces $\ZZ^x$ and $\ZZ^y$ are imposed. If control on these spaces is desired, one could sample from a desired set in the latent space $\tilde{z}_x\in E\subset \ZZ^x$ and add one of the following losses to \eqref{eq:par_min}: 
\begin{equation*}
\alpha_{z_x\aee}D_4\big(\tilde{z}_x,z_x\aee\big)~\text{ or }~\alpha_{z_x\sig}D_5\big(\tilde{z}_x,z_x\sig\big). 
\end{equation*}
This means the sampled latent code is decoded to a signal $x$ (gray in Figure \ref{fig:L-SVD-tikz}), after which it takes either the blue autoencoder path or the red reconstruction path, without the final decoder step $\phi\decx$. Although not guaranteed, it is likely that due to this additional loss, the encoder $\phi\encx$ will map all samples $x_{(i)}$ in the training set to this subset $E$. If this is the case, it means that we have control over the latent space $\ZZ^x$. Moreover it turns out that having a bound on $D_5\big(\tilde{z}_x,z_x\sig\big)$ enables us to compute a uniform error bound for the reconstruction procedure (see Section \ref{sec:analysis_fully}).
\end{itemize}

\section{Analysis}\label{sec:analysis}
In this section, we first consider the case where the encoder and decoder are derived from the SVD of $A$. In Section \ref{sec:Bayesian_IP}, we show that training this L-SVD model with a linear scaling coincides with learning the covariance matrix of a prior in Bayesian inverse problems. In Section \ref{sec:nonl_Tikh}, we provide a convergence analysis with respect to noise in case a nonlinear scaling is used. Finally, in Section \ref{sec:analysis_fully}, we provide a stability and error estimate for the L-SVD model with nonlinear encoding and decoding.

\subsection{Connection with Bayesian inverse problems}\label{sec:Bayesian_IP}
Here we provide an explicit connection between a linear L-SVD model and the solution of a Bayesian inverse problem with Gaussian noise, Gaussian prior and known forward operator $A$. For an introduction to statistical and Bayesian inverse problems, we refer to \cite{Evans2002, Kaipio2005, Stuart2010}.

\subsubsection{Learning the prior covariance matrix}\label{sec:cov_matrix}
\begin{center}
\begin{adjustwidth}{0.04\textwidth}{0.04\textwidth}
\vspace{4mm}
\begin{proposition}\label{prop:bayesian}
Let $x\in \XX\subset \R^m$, $\tilde{y}\in\YY\subset \R^n$ and $\eta\sim\N(0,B)$, where $\XX$ and $\YY$ are Banach spaces. Consider the inverse problem
\begin{equation*}%\label{eq:IP2}
\tilde{y} = Ax +\eta,
\end{equation*}
where $A\in\R^{n\times m}$ has full row-rank, i.e. $\text{rank}(A)=n\leq m$, with thin SVD decomposition $A = US_nV_n^*$. Moreover, let $\mu_0\sim\N(0,C_0)$ be a Gaussian prior measure on $x$. We define $\tilde{B}:=U^*BU$ and restrict the covariance matrix $C_0$ to be of rank $n$ that can be written as $C_0 = V_nC_{\Vn}V_n^*$, where $C_{\Vn}$ is positive definite.\\
$~$\\
Then the maximum a posteriori (MAP) estimate $x_{\text{\tiny MAP}} :=\argmax_x p(y| x)p(x)$ can be written as an SVD inversion method in the following way:
\begin{equation}\label{eq:bayesian_SVD}
\begin{aligned}
x_{\text{\tiny MAP}} &= V_n\Sigma U^*\tilde{y}\\
\text{with }~\Sigma &= \left[\tilde{B}(C_{\Vn}S_n)^{-1}+S_n\right]^{-1}.
\end{aligned}
\end{equation}
\end{proposition}
\vspace{4mm}
\end{adjustwidth}
\end{center}
For the proof we refer to Appendix \ref{app:bayesian_proof}. The connection between \eqref{eq:bayesian_SVD} and L-SVD is clear if we define the linear measurement encoder to be $\phi\ency:=U^*$, the linear signal decoder to be $\phi\decx:=V_n$ and we assume the noise covariance matrix $B$ to be known. Then it can be seen in \eqref{eq:bayesian_SVD} that learning a linear $\Sigma$ is equivalent to optimising over the prior covariance matrix $C_0$, defined via $C_{\Vn}$.

\subsubsection{Scale dependency on Gaussian noise level}
For many inverse problems one assumes an additive noise term that originates from the Gaussian distribution $\eta^\delta\sim\N(0,\dId)$, where the noise level $\delta$ is either known or estimated. If the data covariance matrix $B$ is replaced with $\dId$, \eqref{eq:bayesian_SVD} is simplified to
\begin{align}\label{eq:def_S2}
\Sigma &= \left[\delta(C_{\Vn}S_n)^{-1}+S_n\right]^{-1}.
\end{align}
This implies a stronger regularisation $(C_{\Vn}S_n)^{-1}$ by an increased noise level $\delta$. Thus, by learning the scales $\Sigma$, we learn the prior distribution on $x$ which regularises our inverse problem. For a prior distribution $C_{\Vn} = \gamma\hspace{.3mm}\Id$, it is easily shown that we get the formulation for classical Tikhonov-regularisation \eqref{eq:Tikhonov} back:
\begin{align*}
\Sigma = \left[\delta(\gamma\hspace{.3mm}\Id S_n)^{-1}+S_n\right]^{-1} = \left[\big(\delta/\gamma + S_n^2\big) S_n^{-1}\right]^{-1} = S_\alpha^{-1}\text{~~for~~}\alpha:=\delta/\gamma.
\end{align*}

\subsection{Data-driven Tikhonov regularisation with L-SVD}\label{sec:nonl_Tikh}
In case the forward operator $A$ is known, it is possible to use the linear SVD encoder and decoder and train a nonlinear scaling $\Sigma$. By doing this in a structured manner, a data-driven Tikhonov regularisation can be learned, where the scales $\Sigma$ nonlinearly depend on the measurements $y^\delta$, potentially via $z_y=U^*y^\delta$.
\vspace{4mm}
\begin{center}
\begin{adjustwidth}{0.04\textwidth}{0.04\textwidth} 
\begin{definition}[Data-driven Tikhonov regularisation]
Let $x\in \XX\subset \R^m$ and $y^\delta\in\YY\subset \R^n$, where $\XX$ and $\YY$ are Hilbert spaces. Let $\NN:\R^k\to\R^k$ be a nonlinear function s.t. for all $y\in\YY$: $0<c_\text{min}\leq\NN(y)\leq C_\text{max}<\infty$. Let $A=USV^*$ be the singular value decomposition as defined in Section \ref{sec:IP_SVD}. We define
\begin{equation}\label{eq:Tikhonov_nonl}
\begin{aligned}
x_\alpha^\delta :&= (A^*A+\alpha\NN(y^\delta))^{-1}A^*y^\delta \\ &= V\big(S^2+\alpha\NN (y^\delta)\big)^{-1}S U^* y^\delta. 
\end{aligned}
\end{equation}
\end{definition}
\end{adjustwidth}
\end{center}
L-SVD provides a direct reconstruction through the encoder, scaling and decoder. The function $\NN$ is a trained neural network that can be bounded by construction. In case it has the form of \eqref{eq:Tikhonov_nonl}, the acquired solution is the unique minimiser of a convex functional:
\vspace{4mm}
\begin{center}
\begin{adjustwidth}{0.04\textwidth}{0.04\textwidth} 
\begin{theorem}[Minimising functional]\label{thm:minimiser}
Let $x_\alpha^\delta$ be defined as in \eqref{eq:Tikhonov_nonl}. Then $x_\alpha^\delta$ is the unique minimiser of the functional 
\begin{equation}\label{eq:func_min}
J_\alpha(x) := \norm{Ax-y^\delta}^2_{}+\alpha\NN(y^\delta)\norm{x}^2_{}
\end{equation}
\end{theorem}
\end{adjustwidth}
\end{center}
\vspace{4mm}
\begin{center}
\begin{adjustwidth}{0.04\textwidth}{0.04\textwidth} 
\begin{proof}
Since $0<c_\text{min}\leq\NN(y)\leq C_\text{max}<\infty$, for $\alpha>0$, $J_\alpha$ is strictly convex. Moreover, $\lim_{\norm{x}_{}\to\infty}J_\alpha(x) = \infty$. Hence, $J_\alpha$ has a unique minimiser, which can be found by checking the first-order optimality condition. This yields the expression of \eqref{eq:Tikhonov_nonl}.
\end{proof}
\end{adjustwidth}
\end{center}
\vspace{4mm}
L-SVD in the form of \eqref{eq:Tikhonov_nonl} provides a regularisation method: in the noisy case $\NN (y^\delta)$ determines which scales should be regularised more and which less. In the limit of noise decreasing to zero, the solution converges to the unregularised solution, as shown in Theorem \ref{thm:conv}. 
\vspace{4mm}
\begin{center}
\begin{adjustwidth}{0.04\textwidth}{0.04\textwidth} 
\begin{theorem}[Convergence]\label{thm:conv}
Let $y\in\text{\normalfont Im}(A)$, $\norm{y-y^\delta}_{}\leq\delta$ and $x_\alpha^\delta$ defined as in \eqref{eq:Tikhonov_nonl}. If $\alpha=\alpha(\delta)$ such that
\begin{equation}\label{eq:alpha_lim}
\lim_{\delta\to0}\alpha(\delta) = 0~~\text{ and }~~\lim_{\delta\to0}\frac{\delta^2}{\alpha(\delta)} = 0
\end{equation}
then 
\begin{equation}\label{eq:limit}
\lim_{\delta\to0} x_\alpha^\delta = A^\dag y
\end{equation}
\end{theorem}
\end{adjustwidth}
\end{center}
\vspace{4mm}
\begin{center}
\begin{adjustwidth}{0.04\textwidth}{0.04\textwidth} 
\begin{proof}
Define the sequence $\{\delta_n\}$ such that $\delta_n\to0$, $\alpha_n:=\alpha(\delta_n)$, $y_n:=y^{\delta_n}$, $x_n:=x_{\alpha_n}^{\delta_n}$. With $J_n$ we define the functional \eqref{eq:func_min} with variables as defined above. By Theorem \ref{thm:minimiser}, $x_n$ is the unique minimiser of $J_n$. Hence with $x^\dag:=A^\dag y$, 
\begin{align*}
\alpha_n \NN(y_n)\norm{x_n}_{}^2\leq J_n(x_n)&\leq J_n(x^\dag)\\
&= \norm{Ax^\dag-y_n}_{}^2+\alpha_n\NN(y_n)\norm{x^\dag}^2_{}\\
&= \norm{AA^\dag y-y_n}_{}^2+\alpha_n\NN(y_n)\norm{x^\dag}^2_{}\\
&\leq \delta_n^2+\alpha_n \NN(y_n)\norm{x^\dag}_{}^2.
\end{align*}
From this, we obtain
\begin{align}\label{eq:bound2}
\norm{x_n}_{}^2\leq&\frac{\delta_n^2}{\alpha_n\NN(y_n)}+\norm{x^\dag}^2_{}\leq\frac{\delta_n^2}{c_\text{min}\alpha_n}+\norm{x^\dag}^2_{}.
\end{align}
Because $x_n$ is bounded, it has a convergent subsequence
\begin{equation*}
x_{n_k}\to v\in\XX.
\end{equation*}
Since the bounded linear operator $A$ is sequentially continuous, 
\begin{equation}\label{eq:seq_cont_A}
Ax_{n_k}\to Av\in\YY.
\end{equation}
Again by Theorem \ref{thm:minimiser}, we obtain that 
\begin{align}
\norm{Ax_{n_k}-y_{n_k}}_{}^2 \leq J_{n_k}(x_{n_k})\leq \delta_{n_k}^2+\alpha_{n_k} C_\text{max}\norm{x^\dag}_{}^2\to 0 \text{ as } k\to\infty.
\end{align}
Together with \eqref{eq:seq_cont_A} this implies
\begin{equation}\label{eq:Azy}
Av = y.
\end{equation}
Since any minimiser of $J_n$ is in $\text{ker}(A)^\perp$, $x_n\in\text{ker}(A)^\perp$ for all $x_n$ and therefore $v\in\text{ker}(A)^\perp$. Together with \eqref{eq:Azy}, by \cite[Theorem 2.5]{Engl1996}, this implies that $v=x^\dag$, so that $x_{n_k}\to x^\dag$. By applying the same argument to all subsequences, we obtain 
\begin{equation}\label{eq:xnx}
x_n\to x^\dag.
\end{equation}
%Now assume there is an $\varepsilon>0$ and a subsequence $\{x_n\}$ such that for all $k\in\mathbb{N}$, $\norm{x_{n_k}}_{}\leq\norm{x^\dag}_{}-\varepsilon$. Then this subsequence would have a further subsequence converging weakly to a $z$ with $\norm{z}_{}\leq\norm{x^\dag}_{}-\varepsilon$, which contradicts \eqref{eq:xnx}. Hence
%\begin{equation}\label{eq:liminf}
%\liminf_{n\to\infty} \norm{x_n}_{}^2\geq\norm{x^\dag}_{}^2.
%\end{equation}
%From \eqref{eq:alpha_lim} and \eqref{eq:bound2} we obtain
%\begin{equation}\label{eq:limsup}
%\limsup_{n\to\infty} \norm{x_n}_{}^2\leq\norm{x^\dag}_{}^2.
%\end{equation}
%Together, \eqref{eq:xnx}, \eqref{eq:liminf} and \eqref{eq:limsup} imply that $x_n\to x^\dag$.
Since the sequence $\{\delta_n\}$ is arbitrarily chosen s.t. $\delta_n\to0$, the desired expression \eqref{eq:limit} follows.
\end{proof}
\end{adjustwidth}
\end{center}
\vspace{4mm}
Finally, the convergence rate of \eqref{eq:Tikhonov_nonl} is derived. The proof of Theorem \ref{thm:conv_rate} makes use of several theorems in \cite{Engl1996}.
\vspace{4mm}
\begin{center}
\begin{adjustwidth}{0.04\textwidth}{0.04\textwidth} 
\begin{theorem}[Convergence rate]\label{thm:conv_rate}
Let $w\in\XX$ s.t. $\norm{w}_{}^2\leq\rho$ and define $y:=Aw$, $x^\dag:=A^\dag y = A^\dag Aw$. Then, the optimal parameter choice is $\alpha\sim\big(\frac{\delta}{\rho}\big)^\frac{2}{3}$, which provides the convergence rate
\begin{equation}\label{eq:conv_rate}
\norm{x_\alpha^\delta-x^\dag}_{}^2 = \mathcal{O}(\delta^\frac{2}{3}).
\end{equation}
\end{theorem}
\end{adjustwidth}
\end{center}
\vspace{4mm}
\begin{center}
\begin{adjustwidth}{0.04\textwidth}{0.04\textwidth} 
\begin{proof}
Let $g_{\alpha,y^\delta}:[0,\norm{A}^2]\to\R$ be defined 
\begin{equation}\label{eq:g}
g_{\alpha,y^\delta}(\lambda) := \frac{1}{\lambda+\alpha\NN_\alpha(y^\delta)}.
\end{equation}
This function meets the assumptions of \cite[Theorem 4.1]{Engl1996}:
\begin{equation*}
|\lambda g_{\alpha,y^\delta}(\lambda)| = \frac{\lambda}{\lambda+\alpha\NN_\alpha(y^\delta)}\leq 1
\end{equation*}
and 
\begin{equation*}
\lim_{\alpha\to0} g_{\alpha,y^\delta}(\lambda) = \frac{1}{\lambda}
\end{equation*}
for all $\lambda\in(0,\norm{A}^2]$. Next, for $\alpha>0$ we define
\begin{equation*}
G_{\alpha,y^\delta}:=\sup_{\lambda\in[0,\norm{A}^2]}|g_{\alpha,y^\delta}(\lambda)| = \sup_{\lambda\in[0,\norm{A}^2]}\frac{1}{\lambda+\alpha\NN_\alpha(y^\delta)} = \frac{1}{\alpha\NN_\alpha(y^\delta)}.
\end{equation*}
Furthermore, we define 
\begin{equation}\label{eq:r}
r_{\alpha,y^\delta}(\lambda):=1-\lambda g_{\alpha,y^\delta}(\lambda)= \frac{\alpha\NN_\alpha(y^\delta)}{\lambda+\alpha\NN_\alpha(y^\delta)}.
\end{equation}
Finally, we define for $0<\mu\leq 1$ 
\begin{equation}\label{eq:omega}
\omega_\mu(\alpha) := \tilde{C}_\text{max}\alpha^\mu, ~~\text{ with } \tilde{C}_\text{max}:=\max\{1,C_\text{max}\}.
\end{equation}
For this $\omega_\mu$, $0<\mu\leq 1$, the requirement in \cite[Theorem 4.3]{Engl1996} holds:
\begin{equation}\label{eq:omega_ineq}
\lambda^\mu|r_{\alpha,y^\delta}| = \frac{\lambda^\mu \alpha\NN_\alpha(y^\delta)}{\lambda + \alpha\NN_\alpha(y^\delta)} \leq (C_\text{max}\alpha)^\mu \leq \omega_\mu(\alpha).
\end{equation}
An expanded derivation of \eqref{eq:omega_ineq} is provided in Appendix \ref{app:omega_ineq}. For $\mu\leq 1$, by \cite[Corollary 4.4]{Engl1996}, the parameter choice
$\alpha\sim\left(\frac{\delta}{\rho}\right)^\frac{2}{2\mu+1}$
is of optimal order in $\{x\in\XX~|~x=(A^\dag A)^\mu w, ~~\norm{w}_{}^2\leq\rho\}$. The best possible convergence rate is obtained with $\mu=1$ for $x^\dag=A^\dag Aw$, $\norm{w}_{}^2\leq\rho$. This provides the convergence rate \eqref{eq:conv_rate}.
\end{proof}
\end{adjustwidth}
\end{center}
\vspace{4mm}
\subsection{Fully learned L-SVD}\label{sec:analysis_fully}
In Section \ref{sec:nonl_Tikh}, it was shown that a data-driven Tikhonov regularisation is obtained by giving L-SVD the structure as shown in \eqref{eq:Tikhonov_nonl}. The linear encoder and decoder are obtained from the SVD of $A$, which allows for the convergence analysis that was provided. In the current section we analyse the more general case where the L-SVD is fully learned, which means that the SVD of $A$ can not be used. Furthermore, we consider the L-SVD with nonlinear encoder and decoder and diagonal matrix $\Sigma$, i.e. the second model choice of Section \ref{sec:variations}. 
\vspace{4mm}
\begin{center}
\begin{adjustwidth}{0.04\textwidth}{0.04\textwidth} 
\begin{definition}[Nonlinear L-SVD with diagonal scaling]\label{def:nonl_LSVD}
Define the nonlinear functions and variables as in Definition \ref{def:LSVD}. Let $\Sigma:\ZZ^y\to \ZZ^x$ be a diagonal matrix $\Sigma\in\R^{k\times k}$ with $\{\sigma_1, \cdots, \sigma_k\}$ on the diagonal.  
\end{definition}
\end{adjustwidth}
\end{center}
\vspace{6mm}
\begin{center}
\begin{adjustwidth}{0.04\textwidth}{0.04\textwidth} 
\begin{theorem}[Stability estimate]
Let the variables and functions in L-SVD be defined as in Definition \ref{def:nonl_LSVD}. Define two different measurements $y^\delta_{(1)}$ and $y^\delta_{(2)}$, s.t. $\norm{y_{(1)}-y_{(2)}}_{\ell^2}\leq\varepsilon_y$ for some $\varepsilon_y\geq0$. Then there is an $M>0$ that depends on the weights and nonlinearities in the L-SVD network such that
\begin{equation}\label{eq:bound_LSVD}
\norm{\hat{x}\sig_{(1)}-\hat{x}\sig_{(2)}}_{\ell^2}\leq M\varepsilon.
\end{equation}
\end{theorem}
\end{adjustwidth}
\end{center}
\vspace{6mm}
\begin{center}
\begin{adjustwidth}{0.04\textwidth}{0.04\textwidth} 
\begin{proof}
From Definition \ref{def:nonl_LSVD} and on its turn Definition \ref{def:LSVD}, we compute the bound
\begin{equation}\label{eq:stability}
\begin{aligned}
\norm{\hat{x}\sig_{(1)}-\hat{x}\sig_{(2)}}_{\ell^2}&\leq\norm{\phi\decx}_\op\norm{\hat{z}_{x,(1)}-\hat{z}_{x,(2)}}_{\ell^2}\\
&\leq |\sigma_\text{max}|~\norm{\phi\decx}_\op \norm{\hat{z}_{y,(1)}-\hat{z}_{f,(2)}}_{\ell^2}\\
&\leq \norm{\phi\ency}_\op|\sigma_\text{max}|~\norm{\phi\decx}_\op \norm{y_{(1)}-y_{(2)}}_{\ell^2}\\
&\leq \norm{\phi\ency}_\op|\sigma_\text{max}|~\norm{\phi\decx}_\op \varepsilon_y=:M\varepsilon_y,
\end{aligned}
\end{equation}
where $\norm{\cdot}_\op$ represents the operator norm. After training, the weights of the L-SVD model are fixed, which provides the desired bound \eqref{eq:bound_LSVD}. 
\end{proof}
\end{adjustwidth}
\end{center}
\vspace{4mm}
\begin{center}
\begin{adjustwidth}{0.04\textwidth}{0.04\textwidth} 
\begin{corollary}
For a nonlinear function 
\begin{equation*}
\phi(x):=\tau(W_L\tau(W_{L-1}\dots\tau(W_1 x)\dots))
\end{equation*}
consisting of $L$ layers with weight matrices $W_l$ and pointwise nonlinearities $\tau(x)$ s.t. $\norm{\tau}_\op=C$, the following bound is achieved: $\norm{\phi}_\op\leq C^L\prod_{l=1}^L\left(\norm{W_l}_{\ell^2}\right)$. In case the nonlinearity is a ReLU, leaky ReLU or hyperbolic tangent, $C=1$ and thus the norm only depends on the norms of the weight matrices. In this case the error estimate \eqref{eq:stability} can be written as
\begin{equation*}
\norm{\hat{x}\sig_{(1)}-\hat{x}\sig_{(2)}}_{\ell^2}\leq |\sigma_\text{max}|~\prod_{l=1}^L\left(\norm{{W_{l,\text{enc}}^y}}_{\ell^2}\right) \prod_{l=1}^L\left(\norm{{W_{l,\text{dec}}^x}}_{\ell^2}\right) \norm{y_{(1)}-y_{(2)}}_{\ell^2}. 
\end{equation*}
The error estimate can thus be controlled by assuring small $\ell^2$-norms of the weight matrices, which can be achieved by adding weight regularisation in the neural network objective function.
\end{corollary}
\end{adjustwidth}
\end{center}
\vspace{6mm}
Next, we prove that an error estimate on the difference between any reconstructed signal and the true solution exists, provided that its associated measurement maps to a ball in the latent space. Before we provide this error estimate, we first prove a supporting Lemma. 

\begin{center}
\begin{adjustwidth}{0.04\textwidth}{0.04\textwidth}
\vspace{6mm}
\begin{lemma}\label{lem:ball}
Let $z\in\R^k$, let $F:\R^k\to \R^k$ be a continuous function. Assume $\forall z\in B_1$, $\norm{z-F(z)}_{\ell^2}<\varepsilon$ for some given $0\leq\varepsilon<1$. Then $B_{1-\varepsilon}\subset F(B_1)$, where $B_r:=\{z\in\R^k~|~\norm{z}_{\ell^2}\leq r\}$ is the closed ball centered at $0$ with radius $r$.
\end{lemma}
\end{adjustwidth}
\end{center}

\begin{center}
\begin{adjustwidth}{0.04\textwidth}{0.04\textwidth}
\vspace{6mm}
\begin{proof}
Let us first define a scaled function $\tilde{F}:\R^k\to \R^k$, with $\tilde{F}(z):=\frac{1}{1+\varepsilon}F(z)$. For this scaled function, $\tilde{F}(B_1)\subset B_1$.\vspace{2mm}
\\
The closed unit ball $B_1$ is a contractible space. By definition of a contractible space \cite{Gamelin1999}: $\exists G: B_1 \times [0,1] \to B_1$ continuous such that for all $z$, $G(z,0) = z_0$ and $G(z,1) = z$, where $z_0$ is the contraction point. Since $\tilde{F}(B_1)\subset B_1$ for all $z\in\tilde{F}(B_1)$, it holds that $G(\tilde{F}(z),0) = z_0$ and $G(\tilde{F}(z),1) = \tilde{F}(z)$. Since both $G$ and $\tilde{F}$ are continuous, its composition is continuous. From this it follows that $\tilde{F}(B_1)$ is a contractible space, which implies that $F(B_1)$ is a contractible space. In other words, $F(B_1)$ does not have any `holes'.\vspace{3mm}
\\
Left to show is that the boundary of $F(B_1)$ lies outside $B_{1-\varepsilon}$, which implies $B_{1-\varepsilon}\subset F(B_1)$, because $F(B_1)$ is contractible. For this we make use of \cite[Theorem 4.22]{Rudin1976}: since $F$ is a continuous mapping of a metric space $(\R^k,\norm{}_{\ell^2})$ into a metric space $(\R^k,\norm{}_{\ell^2})$, and the boundary of the unit ball (i.e. $\dd B_1$) is a connected subset of $\R^k$, this implies $F(\dd B_1)$ is connected. Moreover, $\forall z\in \dd B_1$, $F(z)\in B_{1+\varepsilon}\backslash B_{1-\varepsilon}$. This implies that the boundary of the unit ball lies completely outside $B_{1-\varepsilon}$, which completes the proof.
\end{proof}
\end{adjustwidth}
\end{center}
\vspace{6mm}
\begin{center}
\begin{adjustwidth}{0.04\textwidth}{0.04\textwidth}
\begin{theorem}[Reconstruction error estimate] Let the variables and functions in L-SVD be defined as in Definition \ref{def:nonl_LSVD}. Assume that for some $0<\varepsilon_z<1$, for all $\tilde{z}_x\in B_1,~\norm{\tilde{z}_x-z_x\sig}_{\ell^2}<\varepsilon_z$. Then there is an $M>0$ that depends on the weights and nonlinearities in the L-SVD network such that for all $\tilde{z}_x\in B_1$, $\norm{\phi\decx(\tilde{z}_x)-\phi\decx(\Sigma\phi\ency(A\phi\decx(\tilde{z}_x)))}_{\ell^2}<M\varepsilon_z$. Moreover, for all $x\in\R^n$ for which $\phi\encx(x)\in B_{1-\varepsilon_z}$, we have the error estimate $\norm{x-\phi\decx(\Sigma\phi\ency(Ax))}_{\ell^2}<M\varepsilon_z$. \end{theorem}
\end{adjustwidth}
\end{center}
\vspace{4mm}
\begin{center}
\begin{adjustwidth}{0.04\textwidth}{0.04\textwidth}
\begin{proof}
The first part of the proof can be obtained by combining the operator norm of the decoding function in $x$ and the given bound:
\begin{align}\label{eq:bound}
&\norm{\phi\decx(\tilde{z}_x)-\phi\decx(\Sigma\phi\ency(A\phi\decx(\tilde{z}_x)))}_{\ell^2}\\
\leq~&\norm{\phi\decx}_\op\norm{\tilde{z}_x-\Sigma\phi\ency(A\phi\decx(\tilde{z}_x))}_{\ell^2}\nonumber\\
<~&\norm{\phi\decx}_\op\varepsilon_z=:M\varepsilon_z\nonumber
\end{align}
For the second part of the proof, we make use of Lemma \ref{lem:ball}. We define $F(\tilde{z}_x):=\Sigma\phi\ency(A\phi\decx(\tilde{z}_x))$, which is a continuous function from $\R^k$ to $\R^k$. Since for all $\tilde{z}_x\in B_1,~\norm{\tilde{z}_x-F(\tilde{z}_x)}<\varepsilon_z$, we know that all elements in the ball $B_{1-\varepsilon_z}$ are in the range of $F(B_1)$. Therefore, for all $z_x\sig\in B_{1-\varepsilon_z}$, there exists a $\tilde{z}_x\in B_1$, such that the same bound \eqref{eq:bound} holds. 
\end{proof}
\end{adjustwidth}
\end{center}
\vspace{6mm}
Note that the reconstruction error estimate depends on $\norm{\phi\decx}_\op$and $\varepsilon_z$. The former can be kept small by regularising the weights of the decoder in the training phase. The latter can be kept small by including $\norm{\tilde{z}_x-\Sigma\phi\ency(A\phi\decx(\tilde{z}_x))}_{\ell^2}$ in the cost function, as described in the last point of Section \ref{sec:variations}. After training, points in the unit ball can be sampled uniformly and passed through the network to examine an actual value for $\varepsilon_z$.
\vspace{4mm}
\section{Research context}\label{sec:literature}
Our paper presents the L-SVD method for solving inverse problems via hybrid autoencoding. Within the method, a low-dimensional (i.e. sparse) representation or manifold is explicitly learned. This method has connections with many research fields, which are pointed out in this section.
\newpage
\subsection{Combining data and models for solving inverse problems}
Recent research in inverse problems seeks to develop a mathematically coherent foundation for combining data-driven deep learning with model-based approaches based on physical-analytical domain knowledge \cite{Arridge2019}. A first class of methods are partially learned variational and iterative methods \cite{Adler2017, Kobler2017, Hauptmann2018b, Boink2019}. These methods can be seen as a learned variant of gradient, proximal or primal-dual methods. They require less iterations than their non-learned counterparts, but the demand on training time is substantial, while the mathematical analysis of these methods is limited. A second approach is to learn an explicit regularisation term \cite{Chen2018, Aggarwal2018, Lunz2018, Li2020}. Signals affected by artefacts are penalised, while the desired signals are not. Reconstructions are of higher quality compared to classical regularisation choices, but their computation time is of the same order. A third approach is to perform learned post-processing of initial reconstructions obtained by classical methods, which may be affected by artefacts \cite{Jin2016, Schwab2018a}. Data-consistent reconstructions can be obtained without an iterative procedure \cite{Schwab2018a}. However, the quality of the reconstructions heavily depends on the initial reconstruction, which is often obtained by applying a pseudo-inverse to the data. 

The above methods depend on precise knowledge of the physical process, modelled by a forward mapping, which is not always available. However, emperically it was shown that learned iterative methods can still be used in case of inexact forward operators \cite{Hauptmann2018b}. A recent work \cite{Lunz2020} aims to improve an inexact forward operator (linear mapping) explicitly with a neural network, after which it is applied in a variational framework. It was proven that small training losses ensure that the optimisation procedure finds a solution close to the one that would have been found with the true forward operator. 

The optimal regularised inverse matrix (ORIM) method \cite{Chung2014, Chung2015} is a data-driven method that aims to find a linear inverse matrix for a noisy inverse problem. A global minimiser of the Bayes' risk is found when there is knowledge about the forward operator and about the probability distribution of the signals. A fully learned variant where the forward operator is not known is also available \cite{Chung2014}, albeit at a higher computational cost.

\subsection{Fully learned image reconstruction}
Our work is closely related to the work of Zeng \textit{et al.} \cite{Zeng2017}. In this paper, the task of superresolution is solved by autoencoding patches of both a low- and high-resolution image and finding a nonlinear mapping between them. Gupta \textit{et al}. \cite{Gupta2017} used this method for the task of removing motion blur, which is a specific case of a deconvolution problem. In both cases, a one-layered autoencoder was applied to patches of the distorted image (measurement) and desired image (signal). This is only possible if measurement and signal lie in the same domain and if the forward mapping has very little effect outside the local patch. We consider a more general method that does not work patch-based and therefore does not assume identical domains for measurement and signal. As a result, the forward mapping may have a global behaviour. 

A different fully learned reconstruction method without this restriction is proposed by Zhu \textit{et al.} \cite{Zhu2018}, where the problem of finding a reconstruction from undersampled MRI data is considered. Their neural network consists of three fully-connected layers, followed by three convolutional layers, which maps Fourier measurements directly to the desired image. A joint low-dimensional manifold is learned implicitly, since there is no explicit low-dimensional representation within the network architecture. In our work, an explicit representation of a joint manifold is learned in the form of two linearly connected latent codes. Results in \cite{Zhu2018} show a clear improvement over non-learned state-of-the-art methods for in-vivo data. 
 
These works display the potential for fully learned methods in image reconstruction and inverse problems in general: high quality reconstructions are obtained, while no exact knowledge of underlying physics or specifics of the measurement system is required. 

\subsection{Manifold learning}
Many relevant inverse problems in medicine, engineering, astronomy or geophysics are large-scale in both signal and measurement space. However, seen from a statistical point of view, probability mass concentrates near manifolds of a much lower dimension than the original data spaces \cite{Bengio2013}. To detect linear manifolds, principal component analysis is a suitable and simple method. However, since manifolds for real data are expected to be strongly nonlinear \cite{Bengio2013}, one needs to make use of nonlinear techniques. One of the best known methods that achieves this is kernel PCA \cite{Scholkopf1998}. In this method, data is mapped to a reproducing kernel Hilbert space by applying a nonlinear kernel, after which the linear PCA is applied. Other methods are principal geodesic analysis \cite{Fletcher2004}, which can be applied for Riemannian manifolds and geodesic PCA \cite{Bigot2017}, which acts in a Wasserstein space, which is nonlinear.

Another approach to learn nonlinear manifolds is to use autoencoders, which can also be seen as a generalisation of linear PCA \cite{Hinton2006}. Autoencoders have shown to learn explicit representations of nonlinear manifolds \cite{Rifai2011} and provide better low-dimensional latent code in terms of clustering and reconstruction performance \cite{Hinton2006} than their linear counterpart. For the inverse problems \eqref{eq:IP} considered in our paper, there is an explicit relationship between signals and measurement via the forward mapping that models the physics. This means that signals and measurements share one data manifold that is learned by connecting two autoencoders. 

The observation of a shared manifold, or ``an unknown underlying relationship between two domains'' \cite{Zhu2017}, is also the idea driving the cycle GAN. Unlike in our paper, the goal of the cycle GAN is not to find a unique and supervised one-to-one mapping from one domain to the other, but to identify the shared parameters and add such elements so that the output is realistic in its respective domain. This cycle GAN was applied for inverse problems in different forms \cite{Senouf2019, Sim2019}. In these works, the manifold is implicitly learned, unlike the explicitly learned representation that we study in our paper. 

\subsection{Transfer learning with autoencoders}
Transfer learning is used to exploit similarities between different tasks to share information necessary for both tasks. Representation learning, such as manifold learning, has a strong influence in transfer learning scenarios \cite{Bengio2013} since the learned representation can guide the supervised reconstruction task. Recent research has shown that autoencoders can be used as a regulariser for a supervised training task, such as classification \cite{Zhang2016, Le2018}. Such networks, coined supervised autoencoders (SAE), help to generalise the supervised problem. They are specifically useful in a semi-supervised scenario, where a lot of unsupervised training data is available, but supervised training pairs are scarce.

L-SVD profits from the same regularisation and generalisation effect by attaching two autoencoders to the supervised reconstruction problem. For inverse problems a semi-supervised scenario is also often encountered: imagine a training set of undersampled medical MRI or CT data sets and a set of high-quality reconstructions; not all pairs are available because not all patients have had a fully sampled scan that is needed for a high-quality reconstruction.

\subsection{Model reduction and learning} 
Model reduction is a mathematical and computational field of study that derives low-dimensional models of complex systems \cite{Bui-Thanh2008, Benner2015}. Via projections and decompositions it is possible to represent approximations of large-scale high-fidelity computational models resulting from discretization of partial differential equations. Recent developments focus on data-driven learning of governing equations \cite{Brunton2015,Rudy2017,Raissi2019} and learned model reduction \cite{Qian2019}. 

Our work focuses on learning the inverse map for problems that are often physics-driven. The nonlinear equations or parameters of this map are implicitly learned through the latent representations via autoencoders.

\subsection{Bayesian inversion and sparsity}
The goal of the Bayesian approach to solving inverse problems is to find the posterior measure, given sampled data and a prior measure \cite{Stuart2010}. The posterior measure will contain information about the relative probability of different outputs, given the data. Often the posterior is too complex to recover and the goal is shifted to finding a maximum a posteriori (MAP) estimate. In Section \ref{sec:Bayesian_IP} we showed that a linear variant of L-SVD coincides with learning the covariance matrix of the prior measure.

Yu \textit{et al.} \cite{Yu2012} developed piecewise linear estimates (PLE), a method based on Gaussian mixture models (GMM), which are estimated via the MAP expectation-maximization algorithm (MAP-EM). The method makes use of GMM as prior measures on local patches, which results in a linear reconstruction model for each patch. One could think of this procedure as finding a patchwork of locally linear tangent spaces which approximate a nonlinear manifold. If measurement and signal domain are the same, it can be shown that PLE is equivalent to learning a linear L-SVD method for a group of similar patches.

\section{Experiments and implementation}\label{sec:implementation}
In this section, we explain three simulation experiments: two that demonstrate the contributions as stated in Section \ref{sec:contrib} in a relatively low-resolution scenario and one that shows the application of L-SVD to a biomedical data-set in a higher resolution. The forward operator in all experiments is chosen to be the Radon transform \cite{Kak2002}, which is a nonlocal linear operator that produces an ill-posed inverse problem. We will refer to the measurements as `sinograms' and to the signals as `images' in their respective spaces $\YY$ and $\XX$. In Appendix \ref{sec:imp}, the neural network architectures and their parameter choices are provided, together with details of the training set. 

\subsection{Experiment 1: from model-based to data-driven}\label{sec:exp1}
The goal of this simulation experiment is to demonstrate the first perspective of Section \ref{sec:contrib}: data-driven solution of inverse problems. This is done by comparing classical non-learned methods to learned methods. For fair comparison and clarity, only linear inversion methods are considered.

This experiment makes use of the MNIST data set \cite{Lecun1998}, after rescaling it to $64\times64$ pixels. We apply the Radon transform with 64 uniformly samples angles, i.e. a `full-angle' Radon operator $A$, with a discretisation of 64 for each angular view. Moreover, there is no bottleneck latent space. This results in equally large spaces $\YY=\ZZ^y=\ZZ^x=\XX=\R^{4096}$. For training, the `clean' simulated full-angle measurements are normalised, after which Gaussian noise with a noise level $\delta=0.05$ is added. The following linear reconstruction methods are compared:
\begin{enumerate}[label=(\alph*)]
\item Tikhonov-regularised reconstruction with optimally chosen $\alpha$ (see Section \ref{sec:Tikhonov}).
\item Truncated SVD reconstruction with optimal truncation number $r$ (see Section \ref{sec:TSVD}).
\item Optimal regularised inverse matrix (ORIM) \cite[Theorem 2.1]{Chung2014}.
\item Data-driven Tikhonov regularisation, i.e. $U$ and $V$ are from the SVD, while $\Sigma$ is a learned diagonal matrix with the structure of \eqref{eq:Tikhonov_nonl} (see Section \ref{sec:nonl_Tikh}).
\item Reconstruction from a learned covariance matrix of the prior, i.e. $U$ and $V$ are from the SVD, while $\Sigma$ is a full matrix that is learned (see Section \ref{sec:cov_matrix}).
\item Fully learned L-SVD: nonlinear L-SVD as in Definition \ref{def:nonl_LSVD} (see Section \ref{sec:analysis_fully}). A regular autoencoder is used on the sinogram side, which means that noise should be reconstructed after the sinogram is encoded and decoded.
\end{enumerate}
%Besides comparing the reconstruction quality of above methods, we examine the `dictionary' of elements that is obtained by decoding canonical basis vectors of the latent space: since all reconstruction methods are linear, the attainable reconstructions lie in the span of this dictionary. By this examination, the change from model-based to data-driven methods becomes evident.
%
By training on sinograms with noise levels between $0$ and $0.2$ instead of the aforementioned $0.05$, the effect of noise on the matrix $\Sigma$ is investigated. Instead of training for several noise levels individually, only one training set is created, where each sinogram has a randomly chosen noise level $\delta\in[0,0.2]$. To process this data set adequately, the static scaling matrix $\Sigma$ is exchanged for a noise-aware component that depends on the noisy input data (c.f. Section \ref{sec:variations}). This component is a nonlinear fully-connected network which takes $z_y$ as input and provides the diagonal scales $(\sigma_1,\sigma_2,\dots,\sigma_k)$ as output, which are multiplied with $z_y$.

\subsection{Experiment 2: from linear autoencoding to hybrid nonlinear autoencoding}\label{sec:exp2}
The goal of this simulation experiment is to demonstrate the second perspective of Section \ref{sec:contrib}: nonlinear encoding is more effective than linear encoding; moreover, combining two nonlinear autoencoders has a regularising effect on the reconstruction and gives a more insightful latent representation than one autoencoder. For this experiment, we make use of the fully learned L-SVD. 

This experiment again makes use of the MNIST data set \cite{Lecun1998}, after rescaling it to $64\times64$ pixels. A limited-angle Radon transform of 8 uniformly sampled angles is applied, with a discretisation of 64 for each angular view. The latent space is chosen to have 64 dimensions, which means that it acts as a bottleneck. This results in the spaces $\YY=\R^{256}$, $\ZZ^y=\ZZ^x=\R^{64}$, $\XX=\R^{4096}$, providing a dimensionality reduction of $12.5\%$ and $1.56\%$ compared to sinogram and image space respectively. Gaussian noise with a noise level $\delta=0.05$ is added to the limited-angle measurements. We analyse the following methods:
\begin{enumerate}[label=(\alph*)]
\item linear autoencoder;
\item nonlinear autoencoder;
\item linear L-SVD;
\item nonlinear L-SVD ($\alpha=0$);
\item nonlinear L-SVD.
\end{enumerate}
Here the first two methods are only applied on the image side and not on the sinogram side, meaning that the autoencoder only connects $\XX$ and $\ZZ^x$. As can be seen in Figure \ref{fig:L-SVD-tikz}, L-SVD connects the sinogram side with the image side, where a denoising autoencoder (DAE) is used on the sinogram side. This means that noise should not be reconstructed after decoding to $\YY$, which allows for a dimensionality reduction. The fourth method has the same network structure of nonlinear L-SVD, but without the autoencoders on either side (i.e. $\alpha_y = \alpha_x = 0$ in \eqref{eq:par_min}), which allows us to investigate the regularising effect of the autoencoders. It is obvious that this experiment investigates the second perspective from Section \ref{sec:contrib}, since the transition from linear to nonlinear is made, as well as the transition from a single to hybrid autoencoder.

\noindent Finally, we also compare nonlinear L-SVD for the following three cases:
\begin{enumerate}[label=(\alph*)]
\item All supervised training pairs $(x,y^\delta)$ are available. 
\item Only $10\%$ of the training samples is available, all in pairs $(x,y^\delta)$.
\item All training samples $x$ and $y^\delta$ are available, but only $10\%$ is paired. 
\end{enumerate}
In the third case, if the training data is unpaired, then the encoders and decoders are only trained by the autoencoders. That is, the reconstruction loss $D_1\big(\hat{x}_{(i)}\sig, x_{(i)}\big)$ in \eqref{eq:par_min} is set to zero. We investigate how L-SVD can exploit the semi-supervised case in which additional unpaired training data is available.

\subsection{Experiment 3: L-SVD on human chest CT images}\label{sec:exp3}
The third simulation experiment demonstrates the capacity of L-SVD to reconstruct biomedical images without any knowledge on the forward operator. It exploits the Low-Dose Parallel Beam CT (LoDoPaB) data set \cite{Leuschner2019}, which is based on the LIDC/IDRI data set \cite{Armato2011}. In our paper, we only make use of the high quality CT reconstructions in the LoDoPaB-CT data set that we use as `ground truth' for our setup. The images are scaled to $128\times128$ pixels, after which a Radon transform of 36 uniformly sampled angles is applied with a discretisation of 192 for each angular view. The latent space of L-SVD is chosen to have 2048 dimensions. This results in the spaces $\YY=\R^{6912}$, $\ZZ^y=\ZZ^x=\R^{2048}$, $\XX=\R^{16384}$, providing a dimensionality reduction of around $29.6\%$ and $12.5\%$ compared to sinogram and image space respectively. Gaussian noise with an signal-to-noise ratio (SNR) of 40dB is added to the measurements. The following reconstruction methods are compared:\newpage
\begin{enumerate}[label=(\alph*)]
\item T-SVD with optimal truncation number $k\leq2048$;
\item optimal regularised inverse matrix (ORIM) \cite{Chung2014};
\item total variation regularisation with optimal regularisation parameter $\alpha$;
\item fully learned nonlinear L-SVD.
\end{enumerate}
Truncated SVD can be seen as the linear counterpart of L-SVD and makes use of the forward operator $A$. To obtain at least the same dimensionality reduction as L-SVD, the truncation number $r$ of T-SVD is chosen smaller than 2048, but such that it gives the best reconstruction quality. ORIM is a linear data-driven method, for which we use the same training set as for L-SVD. For computational reasons, we make use of \cite[equation (12)]{Chung2014}: the ORIM method that requires knowledge about the mean and covariance of signals $x$ and noise $\eta^\delta$ (see \eqref{eq:IP}). Moreover, the forward operator $A$ is needed for this method. Total variation (TV) \cite{Rudin1992} is an adequate nonlinear regularised reconstruction method for this problem, since the ground truth CT-images consist of almost entirely piecewise constant structures. This method also needs the forward operator $A$: we minimise the functional $\min_x\norm{Ax-y^\delta}+\alpha\text{TV}(x)$, implemented as described in \cite{Boink2018}. 

\section{Numerical results}\label{sec:results}
In this section, the results of the simulation experiments explained in Section \ref{sec:implementation} are shown and discussed. 

\subsection{Experiment 1: from model-based to data-driven}\label{sec:results_exp1}
\begin{figure}[!ht]
\centering
\begin{subfigure}[t]{0.24\textwidth}
\centering
\captionsetup{justification=centering}
\scriptsize{\hspace{3cm}}
\includegraphics[width=0.7\textwidth]{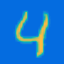}
\caption{ground truth}
\label{fig:LSVD-0}
\end{subfigure}
\begin{subfigure}[t]{0.24\textwidth}
\centering
\captionsetup{justification=centering}
\scriptsize{PSNR: $19.45\pm0.80$}
\includegraphics[width=0.7\textwidth]{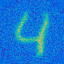}
\caption{classic Tikhonov}
\label{fig:LSVD-1}
\end{subfigure}
\begin{subfigure}[t]{0.24\textwidth}
\centering
\scriptsize{PSNR: $26.81\pm0.95$}
\captionsetup{justification=centering}
\includegraphics[width=0.7\textwidth]{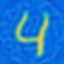}
\caption{T-SVD}
\label{fig:LSVD-2}
\end{subfigure}
\begin{subfigure}[t]{0.24\textwidth}
\centering
\captionsetup{justification=centering}
\scriptsize{PSNR: $27.68\pm0.93$}
\includegraphics[width=0.7\textwidth]{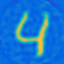}
\caption{\scalebox{0.95}[1.0]{data-driven Tikhonov}\\(diagonal $\Sigma$, fixed $U,V$)}
\label{fig:LSVD-4}
\end{subfigure}\\
\begin{subfigure}[t]{0.24\textwidth}
\centering
\captionsetup{justification=centering}
\scriptsize{PSNR: $28.65\pm0.98$}
\includegraphics[width=0.7\textwidth]{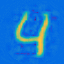}
\caption{full $\Sigma$, fixed $U,V$}
\label{fig:LSVD-6}
\end{subfigure}
\begin{subfigure}[t]{0.24\textwidth}
\centering
\captionsetup{justification=centering}
\scriptsize{PSNR: $29.62\pm1.18$}
\includegraphics[width=0.7\textwidth]{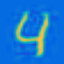}
\caption{ORIM}
\label{fig:LSVD-3}
\end{subfigure}
\begin{subfigure}[t]{0.24\textwidth}
\scriptsize{PSNR: $30.47\pm2.02$}
\centering
\captionsetup{justification=centering}
\includegraphics[width=0.7\textwidth]{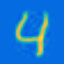}
\caption{fully learned L-SVD \\initialised with SVD}
\label{fig:LSVD-7}
\end{subfigure}
\begin{subfigure}[t]{0.24\textwidth}
\scriptsize{PSNR: $31.08\pm2.28$}
\centering
\captionsetup{justification=centering}
\includegraphics[width=0.7\textwidth]{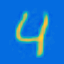}
\caption{fully learned L-SVD \\initialised randomly}
\label{fig:LSVD-8}
\end{subfigure}
\vspace{-4mm}
\caption{Reconstructions $\hat{x}^\Sigma$ (see Figure \ref{fig:L-SVD-tikz}) for different models ranging from fully model-based (top left) to more data-driven (bottom right) with increasing amount of learning.}
\label{fig:LSVD-AE}
\vspace{-3mm}
\end{figure}

The results of the first simulation experiments for a randomly chosen sample in the test set are shown in Figure \ref{fig:LSVD-AE}. For this sample, Gaussian noise with a noise level $\delta=0.05$ was added. Visually, the reconstruction improves gradually as we move from model-based methods with only one tunable parameter (b,c) to combinations of model- and data-driven methods with tunable scaling matrix (d,e) to fully data-driven methods (f,g,h). This is most noticeable in the background of the reconstruction, which should be constant. The visual improvement is confirmed by the peak signal-to-noise ratio (PSNR), displayed above each reconstruction. These values represent the mean and standard deviation of the PSNR values for the first 1000 images in the test set.

\begin{figure}[!ht]
\centering
\begin{subfigure}[t]{0.495\textwidth}
\captionsetup{justification=centering}
\includegraphics[width=\textwidth]{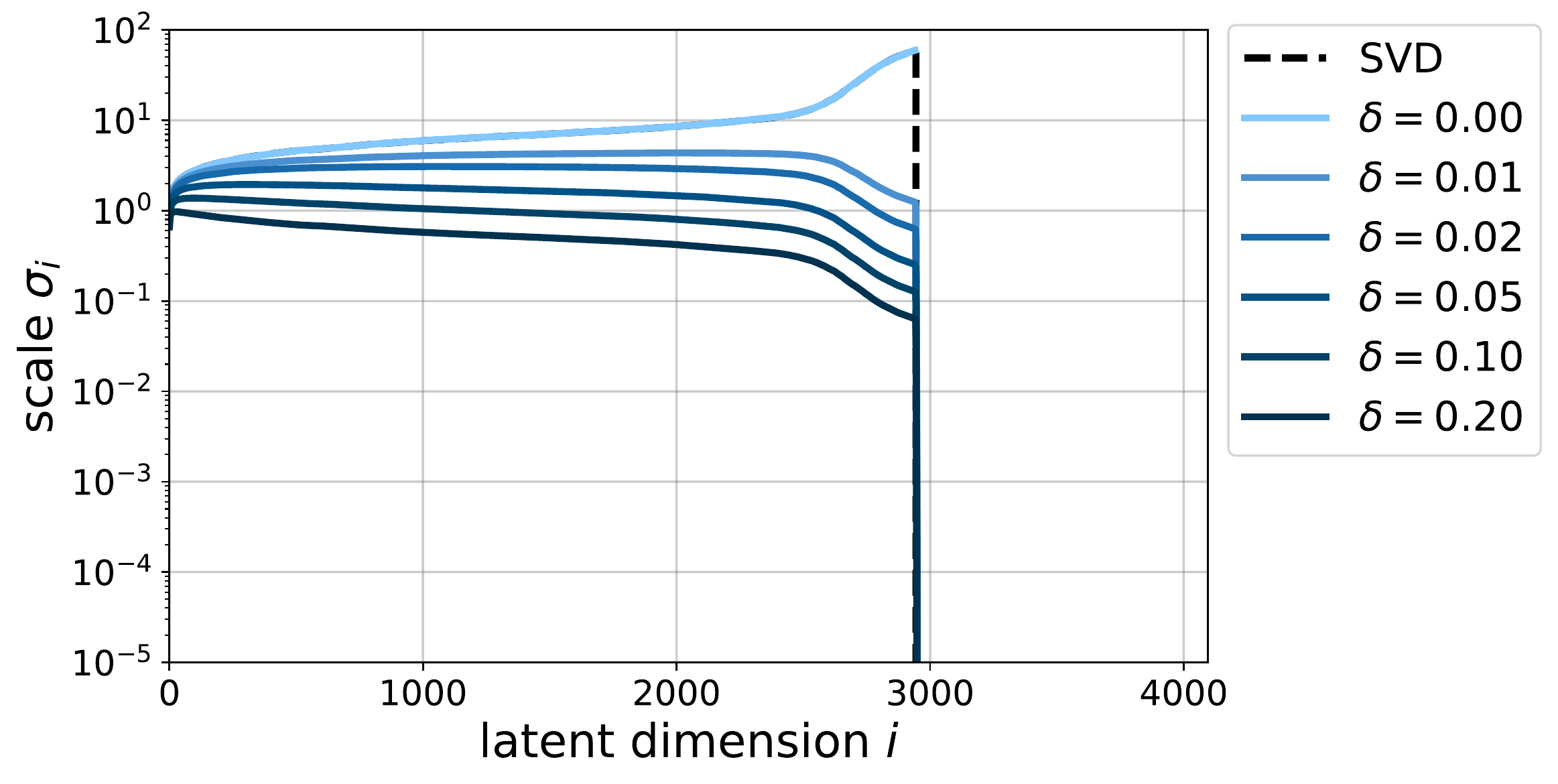}
\caption{Tikhonov regularisation}
\label{fig:graph-Tikhonov}
\end{subfigure}
\hspace{-1mm}
\begin{subfigure}[t]{0.495\textwidth}
\captionsetup{justification=centering}
\includegraphics[width=\textwidth]{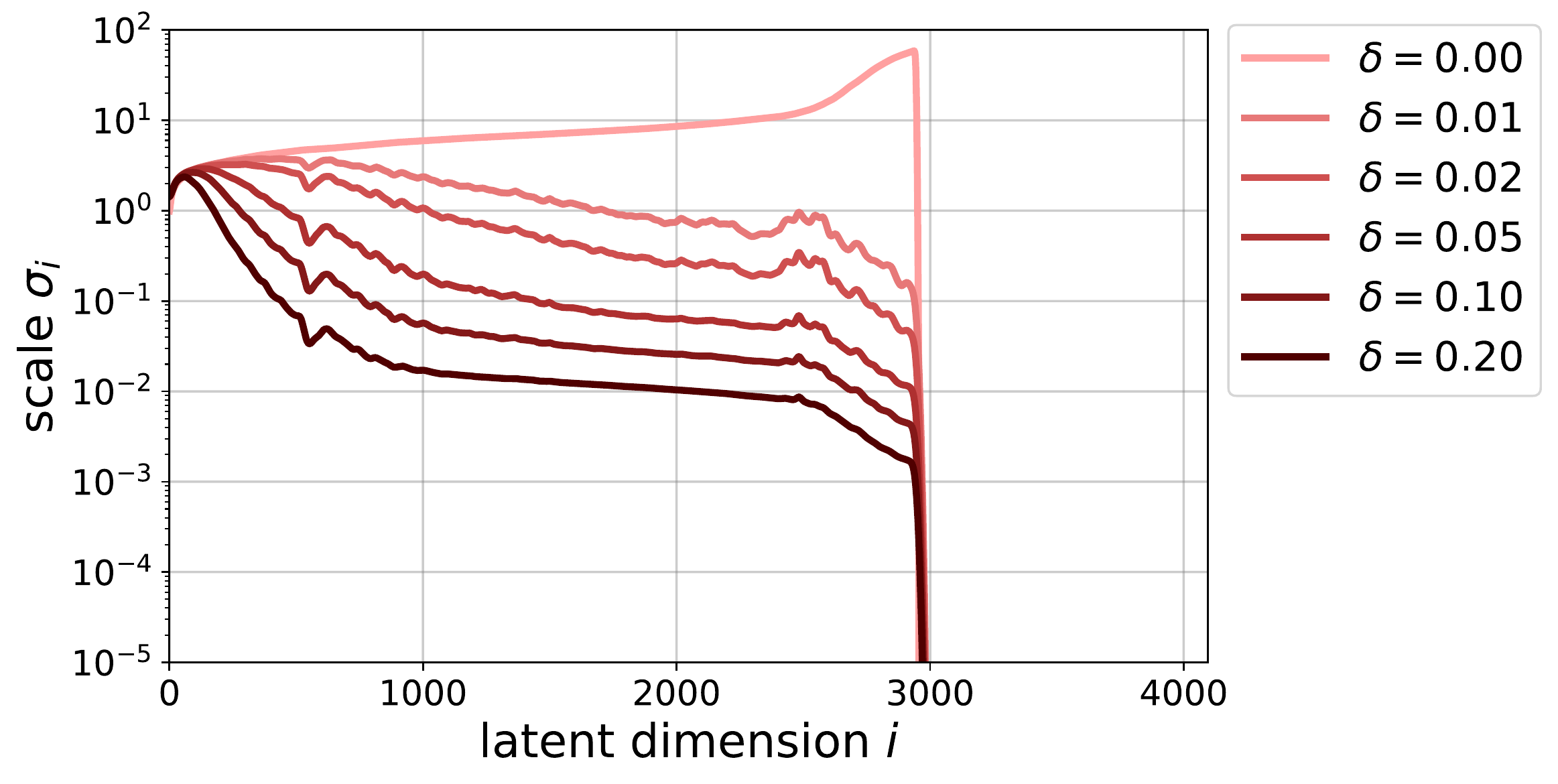}
\caption{Data-driven Tikhonov: $U$,$V$ fixed}
\label{fig:graph-NN-Tikhonov}
\end{subfigure}
\begin{subfigure}[t]{0.495\textwidth}
\captionsetup{justification=centering}
\includegraphics[width=\textwidth]{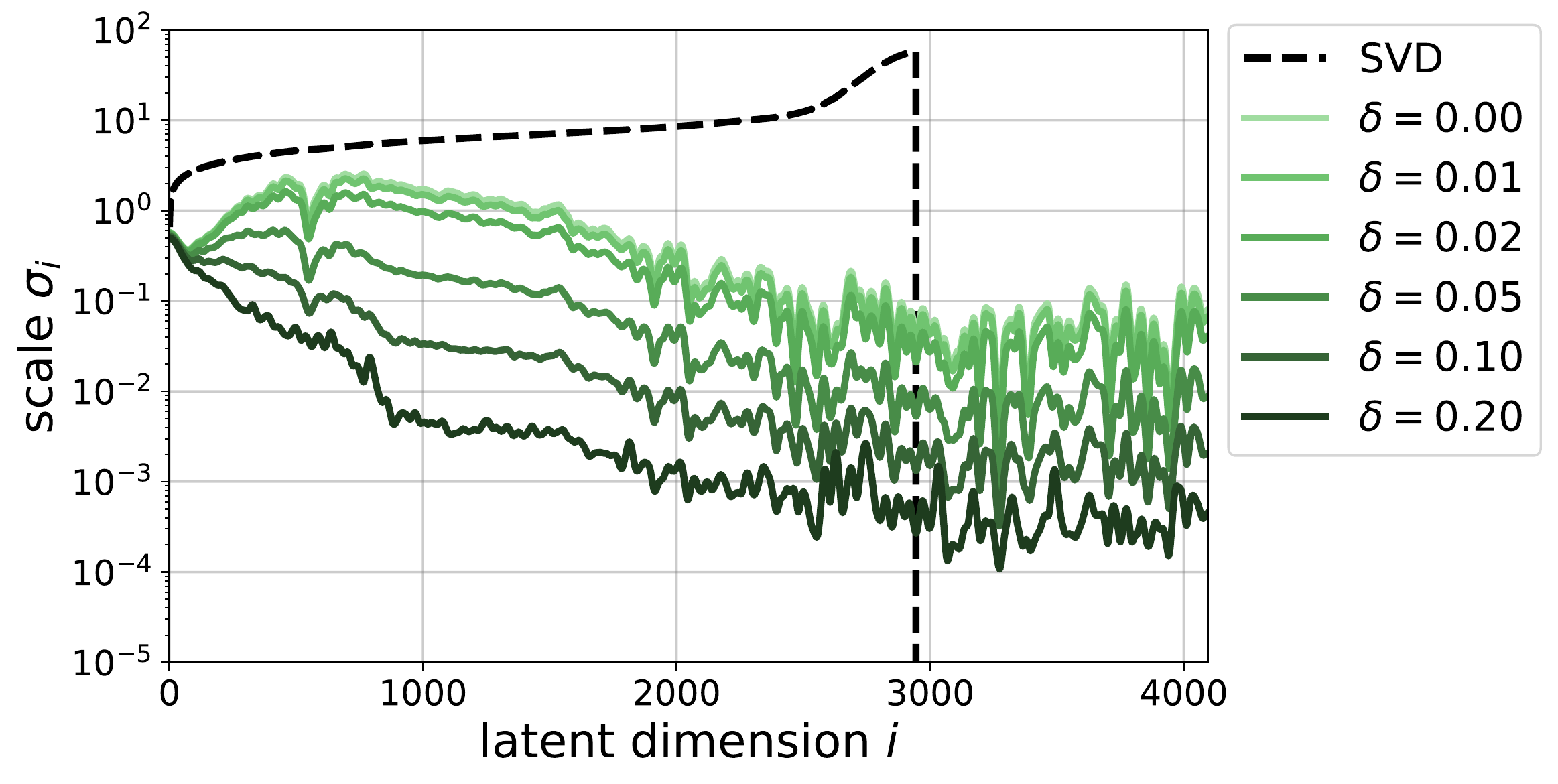}
\caption{fully learned L-SVD initialised with SVD}
\label{fig:graph-SVDinit-x}
\end{subfigure}
\hspace{-1mm}
\begin{subfigure}[t]{0.495\textwidth}
\captionsetup{justification=centering}
\includegraphics[width=\textwidth]{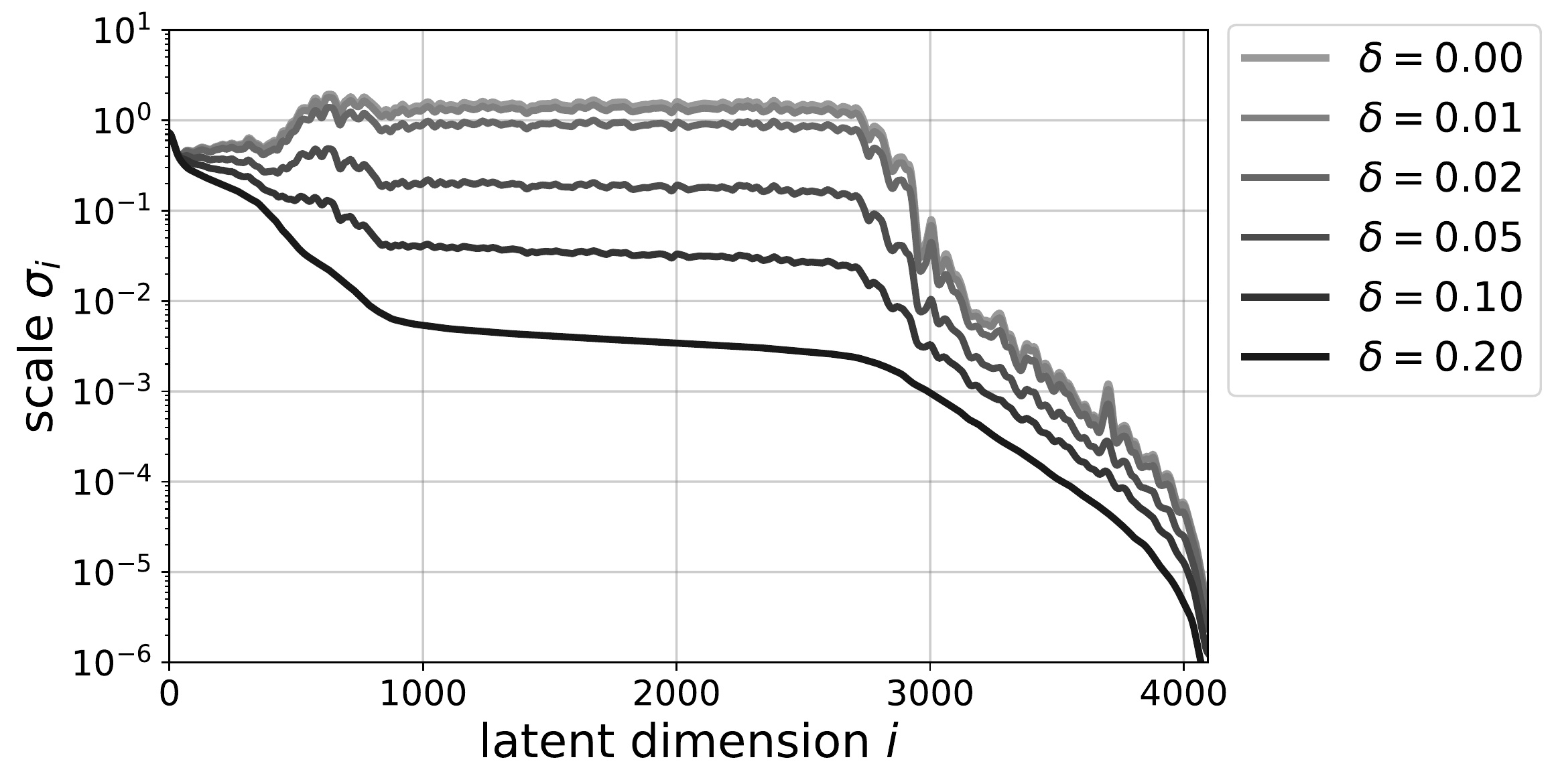}
\caption{fully learned L-SVD initialised randomly}
\label{fig:graph-noSVDinit}
\end{subfigure}
\vspace{-4mm}
\caption{Comparison of the scales $\sigma_i$ depending on noise level for 
methods with increasing amount of learning. Although similar, methods (b-d), where the scales are learned individually, show a greater noise dependency than (a) Tikhonov regularisation. Note that methods (a-c) use the SVD ordering, while (d) is ordered from large to small scales at the highest noise level.}
\label{fig:graph-SVD-noSVD}
\vspace{-6mm}
\end{figure}

%In Table \ref{tab:loss}, the mean squared error (MSE) test losses of all compared methods are shown. For this test set, Gaussian noise with a noise level $\delta = 0.05$ was added. The reconstruction test loss shows that higher quality reconstructions are obtained as the methods become more data-driven. For the fully data-driven method of L-SVD, autoencoder (AE) test losses are much smaller than reconstruction test losses, which is most likely due to the noise in the sinograms that is not added to the AE input. 
%
%\begin{table}[!ht]
%\begin{center}
%\def\arraystretch{1.05}
%\begin{tabular}{ l l l l } 
%\toprule
%Network & Reconstruction 	& AE($x$) 	& AE($y$) \\
%		& test loss			& test loss	& test loss\\	
%\midrule
%Tikhonov 								& 17.4  & 0  	& 0		\\
%truncated SVD							& 6.70  & 3.72  & 4.94	\\
%learned diagonal $\Sigma$, fixed $U,V$ 	& 5.00  & 0  	& 0		\\
%learned full $\Sigma$, fixed $U,V$  	& 3.88  & 0		& 0		\\
%L-SVD initialised with SVD 				& 2.24  & 0.095 & 0.26	\\
%L-SVD initialised randomly   			& 1.93  & 0.058 & 0.25	\\
%\bottomrule
%\end{tabular}
%\end{center}
%\caption {Comparison of test losses (MSE) for different models ranging from fully model-based to more data-driven with increasing amount of learning.}
%\label{tab:loss}
%\vspace{-3mm}
%\end{table}
\newpage
Next, we analyse the effect of noise on the diagonal scaling $\Sigma$ with the network component as explained at the end of Section \ref{sec:exp1}. We compare the networks where encoder and decoder are fixed as $U^*$ and $V$ with the fully learned networks. All networks are compared with Tikhonov regularisation, in which only one tunable parameter $\alpha$ is chosen such that the smallest MSE is obtained. Gaussian noise is added with 6 different noise levels to all sinograms in the test set. The average scales per noise level are shown in Figure \ref{fig:graph-SVD-noSVD}, where (a-c) have the same ordering as the classical SVD and (d) is ordered from large to small scales. For visualisation purposes, the graphs were smoothed by a Gaussian filter with scale 10.

All methods show a decay of scales as the noise level grows. Tikhonov regularisation shows a very similar decay over all latent dimensions, while the decay is much more dimension dependent in (b-d). Here, the scales that coincide with large $s_i$ in the SVD case, i.e. the first dimensions, are relatively noise independent. The scales that coincide with small $s_i$ in the SVD case, i.e. dimensions 500-3000, show a relatively large decay, meaning that they are more affected by noise. Finally, the last part of all graphs show more than a thousand dimensions that have a small scale for all noise levels. To sum up, the behaviour of the scales to noise is similar for all methods in which the scales are learned individually, regardless of the encoding and decoding used. Moreover, most structural information is encoded in a limited number of latent dimensions, while the other dimensions encode of a substantial amount of noise: a compression is learned.

\subsection{Experiment 2: from linear autoencoding to hybrid nonlinear autoencoding}\label{sec:results_exp2}
\begin{figure}[!ht]
\centering
\begin{subfigure}[t]{0.160\textwidth}
\captionsetup{justification=centering}
\includegraphics[width=\textwidth]{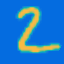}
\caption{\scalebox{0.95}[1.0]{ground truth}}
\label{fig:LSVD-DAE-1}
\end{subfigure}
\begin{subfigure}[t]{0.160\textwidth}
\captionsetup{justification=centering}
\includegraphics[width=\textwidth]{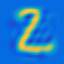}
\caption{linear\\ AE ($\hat{x}\aee$)}
\label{fig:LSVD-DAE-2}
\end{subfigure}
%\begin{subfigure}[t]{0.13\textwidth}
%\captionsetup{justification=centering}
%\includegraphics[width=\textwidth]{LSVD-linear-ae1.png}
%\caption{\\linear\\L-SVD ($\hat{x}\aee$)}
%\label{fig:LSVD-DAE-4}
%\end{subfigure}
\begin{subfigure}[t]{0.160\textwidth}
\captionsetup{justification=centering}
\includegraphics[width=\textwidth]{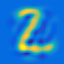}
\caption{linear\\L-SVD ($\hat{x}\sig$)}
\label{fig:LSVD-DAE-6}
\end{subfigure}
\begin{subfigure}[t]{0.160\textwidth}
\captionsetup{justification=centering}
\includegraphics[width=\textwidth]{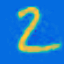}
\caption{nonlinear\\AE ($\hat{x}\aee$)}
\label{fig:LSVD-DAE-3}
\end{subfigure}
\begin{subfigure}[t]{0.160\textwidth}
\captionsetup{justification=centering}
\includegraphics[width=\textwidth]{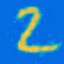}
\caption{nonlinear \\L-SVD$_{\alpha=0}$($\hat{x}\sig$)}
\label{fig:LSVD-DAE-8}
\end{subfigure}
%\begin{subfigure}[t]{0.13\textwidth}
%\captionsetup{justification=centering}
%\includegraphics[width=\textwidth]{LSVD-nonlinear-ae1.png}
%\caption{\\nonlinear L-SVD ($\hat{x}\aee$)}
%\label{fig:LSVD-DAE-5}
%\end{subfigure}
\begin{subfigure}[t]{0.160\textwidth}
\captionsetup{justification=centering}
\includegraphics[width=\textwidth]{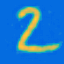}
\caption{nonlinear L-SVD ($\hat{x}\sig$)}
\label{fig:LSVD-DAE-7}
\end{subfigure}
\vspace{-3mm}
\caption{Comparison of the outputs $\hat{x}\aee$ and $\hat{x}^\Sigma$ (see Figure \ref{fig:L-SVD-tikz}) for linear and nonlinear variants of the AE and fully learned L-SVD network.}
\label{fig:LSVD-DAE}
\vspace{-4mm}
\end{figure}

The results of the second simulation experiments for a randomly chosen sample in the test set are shown in Figure \ref{fig:LSVD-DAE}. For the conciseness of this section, only $\hat{x}\sig$ is shown for all variants of L-SVD: not much difference between $\hat{x}\sig$ and $\hat{x}\sig$ was observed. For the autoencoders $\hat{x}\aee$ is shown, since $\hat{x}\sig$ is not available there. Note that for $\hat{x}\aee$, no noise was added to their inputs $x$, while for the reconstruction outputs $\hat{x}\sig$, noise with a noise-level of $0.05$ was added to their inputs $y^\delta$. 

It can be seen that the linear methods produce side-lobes to the main intensities, which are often observed in frequency-based compressions. The nonlinear methods provide a more homogeneous background in the reconstruction, especially L-SVD.

\begin{table}[!ht]
\begin{center}
\def\arraystretch{1.05}
\begin{tabular}{ l l l l l } 
\toprule
Type & Network & Output & Train loss & Test loss \\
\midrule
linear      & AE 					& $\hat{x}\aee$	& 10.2  & 10.1\\
%			& L-SVD 				& $\hat{x}\aee$  & 10.5  & 10.3\\
			& L-SVD 				& $\hat{x}\sig$ & 13.7  & 13.4\\
nonlinear 	& AE 					& $\hat{x}\aee$	& 0.40  & 4.61\\
			& L-SVD$_{\alpha=0}$	& $\hat{x}\sig$	& 1.22  & 5.34\\
%			& L-SVD 				& $\hat{x}\aee$  & 1.49  & 3.97\\
			& L-SVD 				& $\hat{x}\sig$ & 2.00  & 3.94\\
\bottomrule
\end{tabular}
\end{center}
\caption {Generalisation performance analysis by comparing train and test losses (MSE). Nonlinear L-SVD with hybrid autoencoder shows a smaller difference between train and test loss than other nonlinear methods, indicating better generalisation performance.}
\label{tab:train_test}
\vspace{-3mm}
\end{table}

In Table \ref{tab:train_test}, the train and test losses of all compared methods are shown. The first thing that can be seen is that for the linear networks, errors are larger then for the nonlinear networks. The second thing is that there is no significant difference between train and test loss for the linear networks, which indicates that they generalise well. This difference is present in the nonlinear networks, but for L-SVD not to the same extent as for AE en L-SVD$_{\alpha=0}$. From the autoencoder point of view, it seems that L-SVD benefits from the sinogram branch of the network in terms of generalisation. From a reconstruction point of view, L-SVD benefits from the incorporation of the two autoencoders, which contribute to its generalisation capacity in the reconstruction output $\hat{x}\sig$. Finally, since the test loss for nonlinear L-SVD is smaller than for L-SVD$_{\alpha=0}$, we conclude that the two autoencoders act as regularisers for the reconstruction.

Next, canonical basis vectors in the latent space are decoded to the image space, to compose a `dictionary' of elements. While in a linear case all outputs could be reconstructed from these elements in a linear way, this is not true for the nonlinear case. This means that this dictionary only gives a partial view on the decoder. 

\begin{figure}[!ht]
\centering
%\begin{subfigure}[t]{0.22\textwidth}
%\captionsetup{justification=centering}
%\includegraphics[width=\textwidth]{dictionary-linear-AE.png}
%\caption{linear AE}
%\label{fig:dictionary-1}
%\end{subfigure}
%\hspace{0.025\textwidth}
\begin{subfigure}[t]{0.23\textwidth}
\captionsetup{justification=centering}
\includegraphics[width=\textwidth]{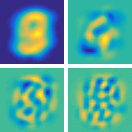}
\caption{linear L-SVD}
\vspace{-3mm}
\label{fig:dictionary-2}
\end{subfigure}
\hspace{0.012\textwidth}
\begin{subfigure}[t]{0.23\textwidth}
\captionsetup{justification=centering}
\includegraphics[width=\textwidth]{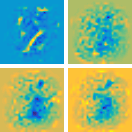}
\caption{nonlinear AE}
\vspace{-3mm}
\label{fig:dictionary-3}
\end{subfigure}
\hspace{0.012\textwidth}
\begin{subfigure}[t]{0.23\textwidth}
\captionsetup{justification=centering}
\includegraphics[width=\textwidth]{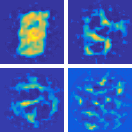}
\caption{\scalebox{0.96}[1.0]{nonlinear L-SVD$_{\alpha=0}$}}
\vspace{-3mm}
\label{fig:dictionary-5}
\end{subfigure}
\hspace{0.012\textwidth}
\begin{subfigure}[t]{0.23\textwidth}
\captionsetup{justification=centering}
\includegraphics[width=\textwidth]{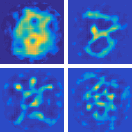}
\caption{nonlinear L-SVD}
\vspace{-3mm}
\label{fig:dictionary-4}
\end{subfigure}
\caption{Selected elements in the latent space $\ZZ^x$, decoded to the image space $\XX$. Nonlinear L-SVD$_{\alpha=0}$ and nonlinear L-SVD learn a more interpretable representation than other methods by combining features from sinogram and image space in their `dictionary'. Due to its similarity to linear L-SVD, linear AE is not shown here. All 64 elements are shown in Appendix \ref{app:dictionary}.}
\label{fig:dictionary}
\vspace{-5mm}
\end{figure}

Figure \ref{fig:dictionary} shows four selected elements that are exemplary for the complete dictionaries, which are provided in Appendix \ref{app:dictionary}. Because the elements of linear AE are very similar to linear L-SVD, they are not shown here. In the top left, linear L-SVD shows an element that is very similar to the Euclidean mean of all training samples, while the top right and bottom images show low and high frequency components that are only active in the middle. Nonlinear AE shows much smaller structures in its elements, which do not seem to have a visual coherent structure. The nonlinear networks L-SVD$_{\alpha=0}$ and L-SVD do provide this visually coherent structure, where L-SVD seems to provide somewhat `smoother' and better connected structures than L-SVD$_{\alpha=0}$. Their dictionaries consist of various elements, of which one is similar to the Euclidean mean (top left), some are similar to digits (top right), some that show a combination of line segments (bottom left) and some with high-frequency components (bottom right) which were also visible in the linear dictionary. With this diversity, fully learned L-SVD combines information from images, sinograms and operator.

\begin{table}[!ht]
\begin{center}
\def\arraystretch{1.05}
\begin{tabular}{l l l l} 
\toprule
& 	\#training pairs & \#training samples	& Test loss \\
\midrule
supervised 		&	60 000  			& 60 000 	& 3.94			\\
supervised		&	\textbf{6 000}		& 6 000	 	& \textbf{13.3}	\\
semi-supervised	&	\textbf{6 000}		& 60 000	& \textbf{9.02}	\\
\bottomrule
\end{tabular}
\end{center}
\caption {Comparison of test losses (MSE) for supervised L-SVD and semi-supervised L-SVD, where $90\%$ of the data is only trained by the autoencoders incorporated in L-SVD.}
\label{tab:semi-supervised}
\vspace{-3mm}
\end{table}

Finally, the capacity of L-SVD in a semi-supervised setup is shown in Table \ref{tab:semi-supervised}. Only $10\%$ of the training samples are available in pairs (i.e. supervised), and the other $90\%$ are available, but not in pairs (i.e. unsupervised). Table \ref{tab:semi-supervised} shows that the semi-supervised setup provides a highly increased performance over the supervised case where only the pairs (i.e. $10\%$) are used. This demonstrates the advantage of adding the autoencoders in L-SVD: they help to efficiently encode and decode, even if pairs between data and signal are not available.

\subsection{Experiment 3: L-SVD on human chest CT images}\label{sec:results_exp3}
\begin{figure}[!ht]
\centering
\begin{subfigure}[t]{0.19\textwidth}
\captionsetup{justification=centering}
\centering
\scriptsize{$~$}\\
\includegraphics[width=\textwidth]{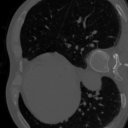}
\caption{ground truth}
\label{fig:Lodopab-1}
\end{subfigure}$\,$
\begin{subfigure}[t]{0.19\textwidth}
\captionsetup{justification=centering}
\centering
\scriptsize{PSNR: 33.62}\\
\includegraphics[width=\textwidth]{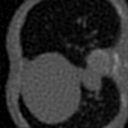}
\caption{T-SVD}
\label{fig:Lodopab-3}
\end{subfigure}$\,$
\begin{subfigure}[t]{0.19\textwidth}
\captionsetup{justification=centering}
\centering
\scriptsize{PSNR: 33.81}\\
\includegraphics[width=\textwidth]{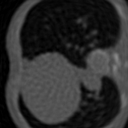}
\caption{ORIM}
\label{fig:Lodopab-35}
\end{subfigure}$\,$
\begin{subfigure}[t]{0.19\textwidth}
\captionsetup{justification=centering}
\centering
\scriptsize{PSNR: 35.23}\\
\includegraphics[width=\textwidth]{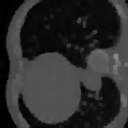}
\caption{TV}
\label{fig:Lodopab-2}
\end{subfigure}$\,$
\begin{subfigure}[t]{0.19\textwidth}
\captionsetup{justification=centering}
\centering
\scriptsize{PSNR: 36.18}\\
\includegraphics[width=\textwidth]{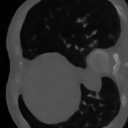}
\caption{L-SVD}
\label{fig:Lodopab-4}
\end{subfigure}
\vspace{-3mm}
\caption{Comparison between fully learned L-SVD and reconstruction methods that make use of the forward operator $A$.}
\label{fig:Lodopab}
\vspace{-4mm}
\end{figure}

Here we demonstrate the capacity of L-SVD to reconstruct biomedical images without any knowledge on the forward operator. We compare it to T-SVD, ORIM and TV, which are all methods that make use of the forward operator $A$. For a randomly chosen sample in the test set, the outputs of all methods are shown in Figure \ref{fig:Lodopab}. It can be seen that none of the methods can reconstruct all details that are apparent in the ground truth. This is probably due to the limited number of angles that were used and the noise added to the sinogram. The T-SVD and ORIM reconstructions give typical ringing artefacts, while the TV reconstruction has the typical staircase behaviour with clusters of piecewise constant structures. The L-SVD reconstruction is smoother than the other methods, which can be seen in the large piecewise constant structure in the bottom left of the image, as well as close to the boundary of the imaged body, where a low contrast between tissues should be reconstructed. To get a better idea of the reconstruction biases of all methods, three additional samples from the test set are provided in Appendix \ref{app:Lodopab}.

In Table \ref{tab:Lodopab}, the peak signal-to-noise ratio (PSNR) and structural similarity (SSIM) of all compared methods are given. 

\begin{table}[!ht]
\begin{center}
\def\arraystretch{1.05}
\begin{tabular}{ l l l } 
\toprule
reconstruction method 	& PSNR 				& SSIM 				\\
\midrule
truncated SVD (T-SVD) 	& $33.86 \pm 1.11$	& $0.845 \pm 0.023$	\\
optimal regularised inverse matrix (ORIM) 	& $34.19 \pm 1.12$	& $0.865 \pm 0.023$	\\
total variation (TV) 	& $35.90 \pm 1.40$	& $0.917 \pm 0.020$	\\
learned SVD (L-SVD) 	& $36.34 \pm 1.62$	& $0.920 \pm 0.023$	\\
\bottomrule
\end{tabular}
\end{center}
\caption {Comparison in PSNR and SSIM between L-SVD and non-learned reconstruction methods that make use of the forward operator $A$.}
\label{tab:Lodopab}
\vspace{-5mm}
\end{table}

Firstly, it can be seen that TV and L-SVD, both nonlinear methods, provide a higher quality than T-SVD and ORIM, linear reconstruction methods. Secondly, from the methods that contain a bottleneck in their architecture, L-SVD is superior to T-SVD. This can be due to its nonlinearity or the fact that it is a learned method. Thirdly, from the data-driven methods, L-SVD shows better results than ORIM, which is probably due to the nonlinearity of L-SVD. Finally, TV and L-SVD are comparable in terms of PSNR and SSIM. However, to use TV as a reconstruction method, it is necessary to know the forward operator $A$ exactly, which is not always possible. Moreover, this quality of TV comes at the expense of computation power: L-SVD is a direct reconstruction method, while TV needs close to a 1000 iterations to converge to its solution. On the other hand, L-SVD requires a large training set and training time, while there is only one parameter to tune for TV.

\section{Conclusion and outlook}\label{sec:conclusion}
We proposed the learned SVD for inverse problems: a reconstruction method that connects low-dimensional representations of corrupted measurements and desired signals. When the forward mapping is known, the L-SVD can be shaped into a data-driven Tikhonov regularisation method, for which we provided a convergence analysis. When the forward mapping is unknown, nonlinear representations of measurements and signals can be fully learned from data, with a connecting layer that is fully connected or sparse, linear or nonlinear, noise dependent or independent. One specific choice is to incorporate the necessary nonlinearity of the learned inverse function in the autoencoders, while the sparse diagonal scaling layer is chosen to be linear, making the connection between measurement and signal manifold easy to understand. In simulation experiments, it was shown that this nonlinear reconstruction gives superior performance to other methods, while providing interpretable autoencoding. Moreover, since the reconstruction error estimate depends on the autoencoding quality, L-SVD can benefit from general advances in nonlinear autoencoding. 

Results show that L-SVD makes use of information from both measurements and signals; by doing so, it learns elements of the physics operator, although not explicitly provided. Therefore the method is especially promising in applications where the forward physics are not completely understood or computationally expensive to simulate. Learning a joint manifold by two connected autoencoders also enables the possibility of a semi-supervised setup: the autoencoders provide regularisation for reconstruction of the signal.

Due to its generic formulation, L-SVD is very flexible for other architecture choices in autoencoding. Therefore, future efforts will lie in investigating other architectures, such as convolutional autoencoders and ladder variational autoencoders \cite{Sonderby2016}, for their inclusion in L-SVD for large scale inverse problems. Furthermore, other loss functions such as the Wasserstein loss or a learned discriminator could be investigated. Finally, we will focus on other ways of incorporating (partial) information of the forward mapping in the architecture of the L-SVD method, to create a regularisation method with a convergence analysis that benefits from the advantages of nonlinear autoencoding.

\section*{Acknowledgements}%
We thank Srirang Manohar, Johannes Schwab, Kathrin Smetana and Leonie Zeune for valuable discussions on the topic. We thank Johannes Schwab for providing his numerical implementation of the Radon transform. The collaboration project is co-funded by the PPP allowance made available by Health$\sim$Holland, Top Sector Life Sciences \& Health, to stimulate public private partnerships. CB acknowledges support by the 4TU programme Precision Medicine and the European Union Horizon 2020 research and innovation programme under the Marie Skodowska-Curie grant agreement No. 777826 NoMADS.

\bibliographystyle{siamplain}
\bibliography{main}

\newpage
\begin{appendices}
\section{Proof of Proposition \ref{prop:bayesian}}\label{app:bayesian_proof}
\begin{center}
\begin{minipage}[c]{0.92\textwidth}
\vspace{4mm}
\begin{proof}

We will first take a look at the more general case where $\mu_0\sim\N(m_0,C_0)$ before we set $m_0=0$. We determine the posterior measure $\mu_\text{post}$ for $x$ given $\tilde{y}$ (see equation (3.4) in \cite{Stuart2010}) as 
\begin{align}\label{eq:Bayes_posterior}
\mu_\text{post} &= \N(m_\text{post},C_\text{post})\nonumber\\
\text{where }~m_\text{post} &= m_0 + C_0A^*(B+AC_0A^*)^{-1}(\tilde{y}-Am_0)\\
\text{and }~C_\text{post} &=C_0 - C_0A^*(B+AC_0A^*)^{-1}AC_0.\nonumber
\end{align}
The maximum a posteriori (MAP) estimate $x_{\text{\tiny MAP}} :=\argmax_x p(y| x)p(x)$ coincides with the mean of $\mu_\text{post}$, i.e. 
\begin{equation*}
x_\text{\tiny MAP} = \argmin_x\left\{\norm{Ax-\tilde{y}}^2_B+\norm{x-m_0}^2_{C_0}\right\} = m_\text{post}.
\end{equation*} 
We substitute $A = US_nV_n^*$ in \eqref{eq:Bayes_posterior}, from which we obtain 
\begin{align*}
m_\text{post} &= m_0+ C_0V_nS_n^*U^*(B+US_nV_n^*C_0V_nS_n^*U^*)^{-1}(\tilde{y}-US_nV_n^*m_0)\\
&= m_0+ C_0V_nS_n^*U^*(U\tilde{B}U^*+US_nV_n^*C_0V_nS_n^*U^*)^{-1}(\tilde{y}-US_nV_n^*m_0)\nonumber\\
&= m_0+ C_0V_nS_n^*(\tilde{B}+S_nV_n^*C_0V_nS_n^*)^{-1}(U^*\tilde{y}-S_nV_n^*m_0)\nonumber\\
&= m_0+ C_0V_n(S_n^{-1}\tilde{B}(S_n^*)^{-1}+V_n^*C_0V_n)^{-1}(S_n^{-1}U^*\tilde{y}-V_n^*m_0)\\
&= m_0+ V_nC_{\Vn}(S_n^{-1}\tilde{B}(S_n^*)^{-1}+C_{\Vn})^{-1}(S_n^{-1}U^*\tilde{y}-V_n^*m_0),
\nonumber
\end{align*}
where we used $\tilde{B}:=U^*BU$ in the second equality and $C_0 = V_nC_{\Vn}V_n^*$ in the last equality. In case of a prior distribution with zero-mean, i.e. $m_0=0$, we get the expression
\begin{align}\label{eq:m_SVD3}
m_\text{post} &= V_n\underbrace{C_{\Vn}(S_n^{-1}\tilde{B}(S_n^*)^{-1}+C_{\Vn})^{-1}S_n^{-1}}_{\Sigma}U^*\tilde{y}.
\end{align}
The symmetric covariance matrix $C_{\Vn}$ is positive definite, hence it is also invertible. We simplify
\begin{align*}\label{eq:def_S}
\Sigma^{-1}&=S_n\big(S_n^{-1}\tilde{B}(S_n^*)^{-1}+C_{\Vn}\big)C_{\Vn}^{-1} \\
&=\tilde{B}(C_{\Vn}S_n)^{-1}+S_n.
\end{align*}
By substituting this expression for $\Sigma$ in \eqref{eq:m_SVD3}, we obtain
\begin{align*}
m_\text{post} &= V_n\Sigma U^*\tilde{y}\nonumber\\
\text{with }~\Sigma &= \left[\tilde{B}(C_{\Vn}S_n)^{-1}+S_n\right]^{-1}.
\end{align*}

\end{proof}
\end{minipage}
\end{center}
\newpage

\section{Inequality \eqref{eq:omega_ineq} written out}\label{app:omega_ineq}
This section shows that the inequality \eqref{eq:omega_ineq} holds, given the \eqref{eq:r} and \eqref{eq:omega}. We consider the function
\begin{equation*}
\lambda^\mu|r_{\alpha,y^\delta}| = \frac{\lambda^\mu \alpha\NN_\alpha(y^\delta)}{\lambda + \alpha\NN_\alpha(y^\delta)}.
\end{equation*}
For $\mu<1$ this function attains its maximum at $\lambda=\big(\mu\alpha\NN_\alpha(y^\delta)\big)/\big(1-\mu\big)$. Filling this in yields 
\begin{align*}
\lambda^\mu|r_{\alpha,y^\delta}| &= \frac{\lambda^\mu \alpha\NN_\alpha(y^\delta)}{\lambda + \alpha\NN_\alpha(y^\delta)} \\
&\leq \frac{ \left(\frac{(\mu\alpha\NN_\alpha(y^\delta))^\mu}{(1-\mu)^\mu}\right)\alpha\NN_\alpha(y^\delta)}{\left(\frac{(\mu\alpha\NN_\alpha(y^\delta))}{(1-\mu)}\right) + \alpha\NN_\alpha(y^\delta)}\\
&= \frac{\big(\mu^\mu(\alpha\NN_\alpha(y^\delta))^{(
\mu+1)}\big)/\big(1-\mu\big)^\mu}{\big(\alpha\NN_\alpha(y^\delta)\big)/\big(1-\mu\big)}\\
&=(1-\mu)^{1-\mu}\mu^\mu\big(\alpha\NN_\alpha(y^\delta)\big)^\mu\\
&\leq \big(\alpha\NN(y^\delta)\big)^\mu \leq (C_\text{max}\alpha)^\mu \leq \tilde{C}_\text{max}\alpha^\mu=\begin{cases}
\alpha^\mu &\text{ if }C_\text{max}<1,\\
C_\text{max}\alpha^\mu &\text{ if }C_\text{max}\geq 1.\end{cases}
\end{align*}
For $\mu=1$ the maximum is attained at $\lambda=\norm{A}^2$ (the right end of the interval for $\lambda$). Filling this in yields
\begin{align*}
\lambda^\mu|r_{\alpha,y^\delta}| &= \frac{\lambda^\mu \alpha\NN_\alpha(y^\delta)}{\lambda + \alpha\NN_\alpha(y^\delta)} \\
&\leq \frac{\norm{A}^{2\mu} \alpha\NN_\alpha(y^\delta)}{\norm{A}^2 + \alpha\NN_\alpha(y^\delta)}\\
&\leq \norm{A}^{2\mu-2} \alpha\NN_\alpha(y^\delta) \\
&= \alpha\NN_\alpha(y^\delta) \leq C_\text{max}\alpha,
\end{align*}
where we filled in $\mu=1$ in the last equality. Combined, this provides the desired inequality \eqref{eq:omega_ineq}.
\newpage

\section{Implementation details}\label{sec:imp}
This appendix provides the implementation details of the neural networks used in this paper. Training is performed in Tensorflow. The first two experiments make use of the complete MNIST training set ($60\,000$ training samples), where each image is rescaled to $64\times64$ using bilinear interpolation. Testing is done on the first 1000 samples of the MNIST test set, using the same interpolation. For these experiments, the Radon transform is applied using the `scikit-image' toolbox in Python. The third experiment makes use of the complete LoDoPaB training set. A data-augmented training set is obtained by mirroring and rotating each image with multiples of $90^\circ$. This yields a training set with $284\,672$ samples: eight times the original number of samples. For this experiment, the Radon transform as described in \cite{Lewitt1990} is applied.

The networks in the first and second experiment only make use of fully-connected (FC) layers. The third experiment has additional convolutional layers on the image side: before the fully-connected layers in the encoder and after the fully-connected layers in the decoder. Details about the encoder and decoder are given in Table \ref{tab:architecture_encdec}. After each layer, except the last layer, a leaky ReLU (lReLU) with parameter $\gamma$ as specified in Table \ref{tab:architecture_encdec} is applied. All networks in the first two experiments are chosen without biases, the third experiment makes use of biases. Initial weights of all experiments are normally distributed with a standard deviation of $0.01$. 

\begin{table}[!ht]
\begin{center}
\begin{tabular}{l l l l} 
\toprule
Parameter 									& Experiment 1f 				& Experiment 2 				& Experiment 3 				\\
\midrule
\# FC layers $\phi\ency$ / $\phi\decy$		& 1 						& 1 						& 1							\\
\# FC layers $\phi\encx$ / $\phi\decx$		& 1 						& 4 						& 3							\\
biases 	& no						& no						& yes						\\
nonlinearity 								& - 		& lReLU$_{\gamma=0.1}$ 		& lReLU$_{\gamma=0.2}$ 		\\
\# conv layers $\phi\ency$ / $\phi\decy$ 	& 0 						& 0 						& 0							\\
\# conv layers $\phi\encx$ / $\phi\decx$ 	& 0 						& 0 						& 3							\\
\# kernels per conv layer					& -				            & - 	 			& 32						\\
kernel size									& - 				        & -				& $5\times5$				\\
\bottomrule
\end{tabular}
\end{center}
\caption{Architecture choices of the encoder and decoder for each of the experiments.}
\label{tab:architecture_encdec}
\vspace{-3mm}
\end{table}
The fully learned L-SVD networks (experiments 1f, 2 and 3) make use of a linear scaling matrix $\Sigma$. Experiments 1d and 1e make use of a nonlinear scaling function $\Sigma(z_y)$ in the form of a neural network, which consists of five fully-connected layers with biases. After each layer a leaky ReLU with parameter $\gamma=0.1$ is applied, except for the last layer: 
\begin{itemize}
\item Experiment 1d applies $\big((C_\text{max}-c_\text{min})\text{sig}(\cdot) + c_\text{min}\big)$ as the final nonlinearity, where $\text{sig}(\cdot)$ denotes the sigmoid function. This nonlinearity ensures the bounds $c_\text{min}\leq\NN(\cdot)\leq C_\text{max}$ (see Section \ref{sec:nonl_Tikh}). In this experiment, $c_\text{min}=10^{-2}$ and $C_\text{max}=10$. The total scaling 
$$\Sigma(z_y) = (\big(S^2+\alpha\NN (z_y)\big)^{-1}S$$ 
is obtained by choosing $\alpha(\delta) = \delta^{\frac{2}{3}}$ and taking $S$ from the SVD of $A$.
\item Experiment 1e applies the softplus function as the final nonlinearity. This immediately yields the total scaling $\Sigma(z_y)$.
\end{itemize}
Initial weights are normally distributed with a standard deviation of $0.01$. 

For all experiments, each loss function $D_j$ in \eqref{eq:par_min} is of $\ell^2$-type (mean squared error) with $\alpha_y,\alpha_x$ as specified in Table \ref{tab:architecture_opt}. We apply the ADAM optimiser using a learning rate with exponential weight decay. The number of epochs, batch sizes and the start and final learning rates are stated in Table \ref{tab:architecture_opt}. All other optimisation parameters are the default choices of ADAM in Tensorflow. Gradient norm clipping with a value of 10 is applied for training stability. No regularisation, dropout or batch normalisation are used. 

\begin{table}[!ht]
\begin{center}
\begin{tabular}{l l l l} 
\toprule
Parameter 									& Experiment 1 				& Experiment 2 				& Experiment 3 				\\
\midrule
loss parameters 							& $\alpha_y=2,~\alpha_x=1$ 	& $\alpha_y=2,~\alpha_x=1$ 	& $\alpha_y=0.2,~\alpha_x=1$\\
\# epochs									& 250 						& 250 						& 500						\\
batch size									& 100 						& 100 						& 100 						\\
start learning rate 						& $10^{-3}$					& $10^{-3}$ 				& $2\cdot10^{-4}$					\\
final learning rate 						& $2\cdot10^{-4}$ 			& $2\cdot10^{-4}$ 			& $2\cdot10^{-5}$					\\
\bottomrule
\end{tabular}
\end{center}
\caption{Optimisation choices for each of the described experiments.}
\label{tab:architecture_opt}
\vspace{-3mm}
\end{table}

\newpage
\section{Visualisation of latent space elements in experiment 1}\label{app:dictionary_linear}
\begin{figure}[!ht]
\centering
\begin{subfigure}[t]{0.45\textwidth}
\captionsetup{justification=centering}
\includegraphics[width=\textwidth]{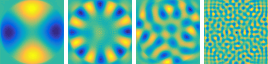}
\caption{SVD decoded to $\XX$}
\label{fig:dictionary-SVD-x}
\end{subfigure}
\hspace{0.025\textwidth}
\begin{subfigure}[t]{0.45\textwidth}
\captionsetup{justification=centering}
\includegraphics[width=\textwidth]{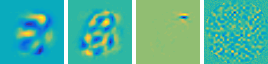}
\caption{L-SVD decoded to $\XX$}
\label{fig:dictionary-noSVDinit-x}
\end{subfigure}
\begin{subfigure}[t]{0.45\textwidth}
\captionsetup{justification=centering}
\includegraphics[width=\textwidth]{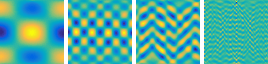}
\caption{SVD decoded to $\YY$}
\label{fig:dictionary-SVD-y}
\end{subfigure}
\hspace{0.025\textwidth}
\begin{subfigure}[t]{0.45\textwidth}
\captionsetup{justification=centering}
\includegraphics[width=\textwidth]{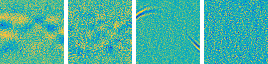}
\caption{L-SVD decoded to $\YY$}
\label{fig:dictionary-noSVDinit-y}
\end{subfigure}\\
\vspace{-3mm}
\caption{Selected elements in the latent space $\ZZ^x$, decoded to the image space $\XX$ and the sinogram space $\YY$. SVD only makes use of the operator, while L-SVD combines operator information with image and sinogram information. This results in more localised information in the decoded elements of the data-driven L-SVD approach.}
\label{fig:dictionary-SVD-noSVD}
\vspace{-4mm}
\end{figure}

To understand the transition from model-based to data-driven, canonical basis vectors in the latent space are decoded to the image space for both the regular SVD (Figure \ref{fig:LSVD-1}-\ref{fig:LSVD-6}) and fully learned L-SVD with random initialisation (Figure \ref{fig:LSVD-8}). Four selected elements from this `dictionary' are shown in Figure \ref{fig:dictionary-SVD-noSVD}.

SVD decomposes the Radon operator in different elements with a different geometrical scale. Moreover, it combines higher order harmonics in the image space and the sinogram space. For example, the second sinogram from the left in Figure \ref{fig:dictionary-SVD-y} shows an approximate 2D sinusoidal structure, while Figure \ref{fig:dictionary-SVD-x} provides its counterpart in the image space. For the third image from the left, it is the other way around. L-SVD shows similar behaviour for larger geometrical scales, but differences are also apparent: the sinusoids are only `active' at the location of potential MNIST digits, and the sinogram space also encodes noise-like structures (first and second image in Figures \ref{fig:dictionary-noSVDinit-x} and \ref{fig:dictionary-noSVDinit-y}). Smaller scale elements (third image) show very localised geometrical structures in image space, while others (fourth image) only seem to capture noise.

\newpage
\section{Visualisation of all latent space elements in experiment 2}\label{app:dictionary}
\begin{figure}[!ht]
\begin{subfigure}[h]{0.11\textwidth}
\captionsetup{justification=centering}
\caption{\\linear \\AE}
\label{fig:dictionary-full-1}
\end{subfigure}
\begin{subfigure}[h]{0.87\textwidth}
\captionsetup{justification=centering}
\includegraphics[width=\textwidth]{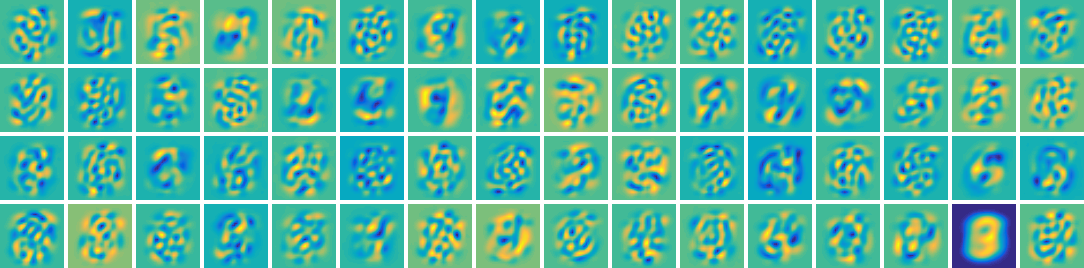}
\end{subfigure}
\vspace{2mm}

\begin{subfigure}[h]{0.11\textwidth}
\captionsetup{justification=centering}
\caption{\\linear L-SVD}
\label{fig:dictionary-full-2}
\end{subfigure}
\begin{subfigure}[h]{0.87\textwidth}
\includegraphics[width=\textwidth]{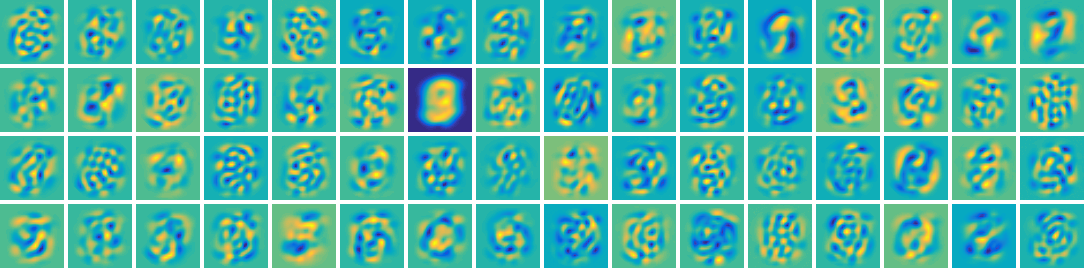}
\end{subfigure}
\vspace{2mm}

\begin{subfigure}[h]{0.11\textwidth}
\captionsetup{justification=centering}
\caption{\\nonlinear \\AE}
\label{fig:dictionary-full-3}
\end{subfigure}
\begin{subfigure}[h]{0.87\textwidth}
\includegraphics[width=\textwidth]{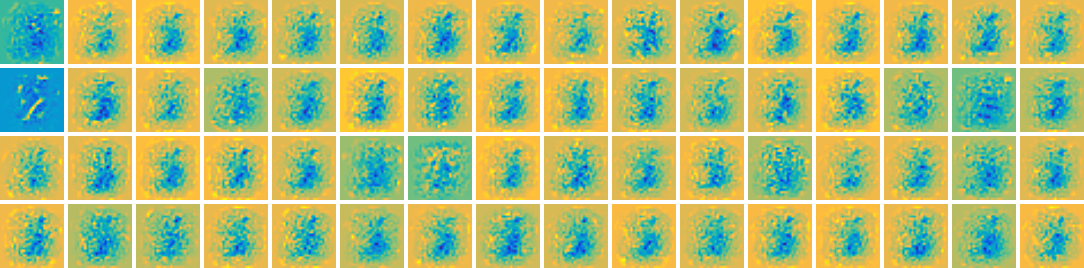}
\end{subfigure}
\vspace{2mm}

\begin{subfigure}[h]{0.11\textwidth}
\captionsetup{justification=centering}
\caption{\\nonlinear \\L-SVD$_{\alpha=0}$}
\label{fig:dictionary-full-4}
\end{subfigure}
\begin{subfigure}[h]{0.87\textwidth}
\includegraphics[width=\textwidth]{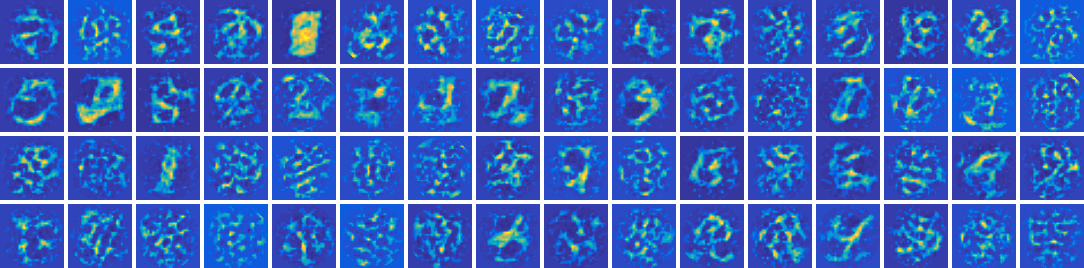}
\end{subfigure}
\vspace{2mm}

\begin{subfigure}[h]{0.11\textwidth}
\captionsetup{justification=centering}
\caption{\\nonlinear \\L-SVD}
\label{fig:dictionary-full-5}
\end{subfigure}
\begin{subfigure}[h]{0.87\textwidth}
\includegraphics[width=\textwidth]{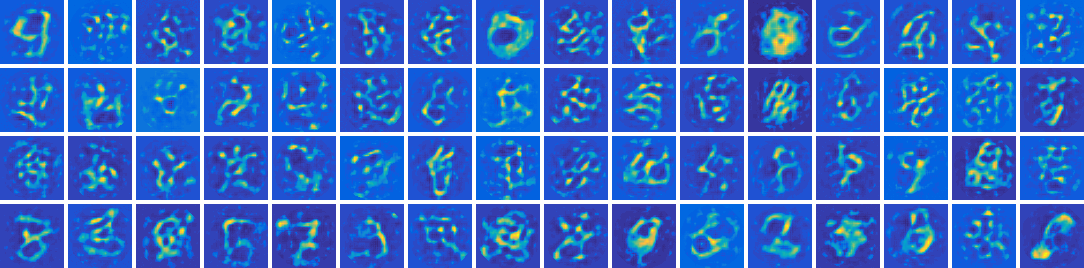}
\end{subfigure}
\vspace{0mm}
\caption{All 64 elements of the dictionary of all methods, from which a selection was shown in Figure \ref{fig:dictionary}.}
\label{fig:dictionary-full}
\vspace{-3mm}
\end{figure}

Figure \ref{fig:dictionary-full} provides the complete dictionaries for all methods, from which a selection was shown in Figure \ref{fig:dictionary}. For a discussion on the results we refer to Section \ref{sec:results_exp2}.
\newpage

\section{Additional results for human chest CT images}\label{app:Lodopab}
$~$
\begin{figure}[!ht]
\centering
\begin{subfigure}[t]{0.19\textwidth}
\captionsetup{justification=centering}
\centering
\scriptsize{$~$}\\
\includegraphics[width=\textwidth]{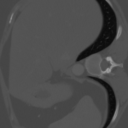}
\caption{ground truth}
\label{fig:App1-Lodopab-1}
\end{subfigure}$\,$
\begin{subfigure}[t]{0.19\textwidth}
\captionsetup{justification=centering}
\centering
\scriptsize{PSNR: 35.23}\\
\includegraphics[width=\textwidth]{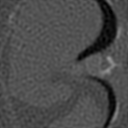}
\caption{T-SVD}
\label{fig:App1-Lodopab-3}
\end{subfigure}$\,$
\begin{subfigure}[t]{0.19\textwidth}
\captionsetup{justification=centering}
\centering
\scriptsize{PSNR: 36.35}\\
\includegraphics[width=\textwidth]{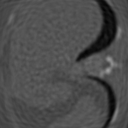}
\caption{ORIM}
\label{fig:App1-Lodopab-35}
\end{subfigure}$\,$
\begin{subfigure}[t]{0.19\textwidth}
\captionsetup{justification=centering}
\centering
\scriptsize{PSNR: 38.97}\\
\includegraphics[width=\textwidth]{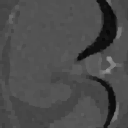}
\caption{TV}
\label{fig:App1-Lodopab-2}
\end{subfigure}$\,$
\begin{subfigure}[t]{0.19\textwidth}
\captionsetup{justification=centering}
\centering
\scriptsize{PSNR: 40.88}\\
\includegraphics[width=\textwidth]{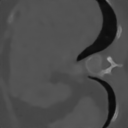}
\caption{L-SVD}
\label{fig:App1-Lodopab-4}
\end{subfigure}
\vspace{-3mm}
\caption{}
\label{fig:App1-Lodopab}
\vspace{-4mm}
\end{figure}

\begin{figure}[!ht]
\centering
\begin{subfigure}[t]{0.19\textwidth}
\captionsetup{justification=centering}
\centering
\scriptsize{$~$}\\
\includegraphics[width=\textwidth]{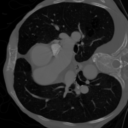}
\caption{ground truth}
\label{fig:App3-Lodopab-1}
\end{subfigure}$\,$
\begin{subfigure}[t]{0.19\textwidth}
\captionsetup{justification=centering}
\centering
\scriptsize{PSNR: 32.88}\\
\includegraphics[width=\textwidth]{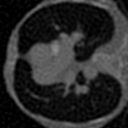}
\caption{T-SVD}
\label{fig:App3-Lodopab-3}
\end{subfigure}$\,$
\begin{subfigure}[t]{0.19\textwidth}
\captionsetup{justification=centering}
\centering
\scriptsize{PSNR: 33.15}\\
\includegraphics[width=\textwidth]{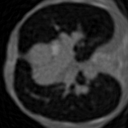}
\caption{ORIM}
\label{fig:App3-Lodopab-35}
\end{subfigure}$\,$
\begin{subfigure}[t]{0.19\textwidth}
\captionsetup{justification=centering}
\centering
\scriptsize{PSNR: 34.54}\\
\includegraphics[width=\textwidth]{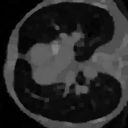}
\caption{TV}
\label{fig:App3-Lodopab-2}
\end{subfigure}$\,$
\begin{subfigure}[t]{0.19\textwidth}
\captionsetup{justification=centering}
\centering
\scriptsize{PSNR: 33.35}\\
\includegraphics[width=\textwidth]{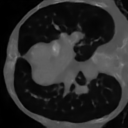}
\caption{L-SVD}
\label{fig:App3-Lodopab-4}
\end{subfigure}
\vspace{-3mm}
\caption{}
\label{fig:App3-Lodopab}
\vspace{-4mm}
\end{figure}

\begin{figure}[!ht]
\centering
\begin{subfigure}[t]{0.19\textwidth}
\captionsetup{justification=centering}
\centering
\scriptsize{$~$}\\
\includegraphics[width=\textwidth]{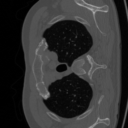}
\caption{ground truth}
\label{fig:App4-Lodopab-1}
\end{subfigure}$\,$
\begin{subfigure}[t]{0.19\textwidth}
\captionsetup{justification=centering}
\centering
\scriptsize{PSNR: 33.69}\\
\includegraphics[width=\textwidth]{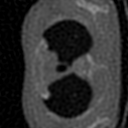}
\caption{T-SVD}
\label{fig:App4-Lodopab-3}
\end{subfigure}$\,$
\begin{subfigure}[t]{0.19\textwidth}
\captionsetup{justification=centering}
\centering
\scriptsize{PSNR: 34.04}\\
\includegraphics[width=\textwidth]{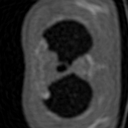}
\caption{ORIM}
\label{fig:App4-Lodopab-35}
\end{subfigure}$\,$
\begin{subfigure}[t]{0.19\textwidth}
\captionsetup{justification=centering}
\centering
\scriptsize{PSNR: 35.33}\\
\includegraphics[width=\textwidth]{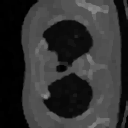}
\caption{TV}
\label{fig:App4-Lodopab-2}
\end{subfigure}$\,$
\begin{subfigure}[t]{0.19\textwidth}
\captionsetup{justification=centering}
\centering
\scriptsize{PSNR: 36.41}\\
\includegraphics[width=\textwidth]{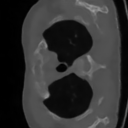}
\caption{L-SVD}
\label{fig:App4-Lodopab-4}
\end{subfigure}
\vspace{-3mm}
\caption{}
\label{fig:App4-Lodopab}
\vspace{-4mm}
\end{figure}

\end{appendices}

\end{document}